%% file: top_arxiv.tex
\documentclass[10pt,twocolumn,letterpaper]{article}

\usepackage{cvpr}
\usepackage{times}
\usepackage{epsfig}
\usepackage{graphicx}
\usepackage{amsmath}
\usepackage{amssymb}

\usepackage[table]{xcolor}
\usepackage{float}
\usepackage{booktabs}
\usepackage{multirow}

\usepackage[pagebackref=true,breaklinks=true,letterpaper=true,colorlinks,bookmarks=false]{hyperref}

\input{defs}

\cvprfinalcopy % *** Uncomment this line for the final submission

\def\cvprPaperID{8458} % *** Enter the CVPR Paper ID here

\ifcvprfinal\fi
\begin{document}

%%%%%%%%% TITLE
%\title{Shape Reconstruction by Learning Continuous Parametric Surface Representation}
\title{Shape Reconstruction by Learning Differentiable Surface Representations}

\author{\vspace{0.5em}
	{Jan Bednarik \quad Shaifali Parashar \quad 
		Erhan Gundogdu \quad Mathieu Salzmann \quad Pascal Fua} \\
	{CVLab, EPFL, Switzerland} \\
	{\small \{firstname.lastname\}@epfl.ch}\\
}

\maketitle
%\thispagestyle{empty}

\input{tex/abstract.tex}
\input{tex/introduction.tex}
\input{tex/related_work.tex}
\input{tex/multi_patch_representations.tex}
\input{tex/differnetiable_properties.tex}
\input{tex/experiments.tex}
\input{tex/conclusion.tex}

{\small
\bibliographystyle{ieee_fullname}
\bibliography{string,graphics,vision,learning,biomed}
}

\input{tex/supplementary_arxiv.tex}

\end{document}

%% file: defs.tex
\newif\ifdraft
\draftfalse
\drafttrue

\definecolor{orange}{rgb}{1,0.5,0}
\definecolor{violet}{RGB}{70,0,170}

\ifdraft
 \newcommand{\PF}[1]{{\color{red}{\bf PF: #1}}}
 
 \newcommand{\MS}[1]{{\color{green}{\bf MS: #1}}}
 
 \newcommand{\JB}[1]{{\color{blue}{\bf JB: #1}}}
 
 \newcommand{\ShP}[1]{{\color{violet}{\bf SP: #1}}}
 
 \newcommand{\ErG}[1]{{\color{orange}{\bf EG: #1}}}
 
\else
 \newcommand{\PF}[1]{}
 
 \newcommand{\MS}[1]{}
 
 \newcommand{\JB}[1]{}
 
 \newcommand{\ShP}[1]{}
 
 \newcommand{\ErG}[1]{}
 
\fi

\newcommand{\comment}[1]{}
\newcommand{\parag}[1]{\vspace{-3mm}\paragraph{#1}}

\newcommand{\real}{\mathbb{R}}
\newcommand{\pb}{\mathbf{p}}

\newcommand{\wb}{\mathbf{w}}
\newcommand{\nb}{\mathbf{n}}
\newcommand{\db}{\mathbf{d}}
\newcommand{\pbi}{\mathbf{p}_{i}}

\newcommand{\qbj}{\mathbf{q}_{j}}
\newcommand{\rb}{\mathbf{r}}
\newcommand{\rbi}{\mathbf{r}_{i}}
\newcommand{\domf}{\mathcal{D}_{f}}
\newcommand{\surf}{\mathcal{S}}
\newcommand{\jac}{\mathbf{J}}
\newcommand{\fu}{\mathbf{f}_{u}}
\newcommand{\fv}{\mathbf{f}_{v}}
\newcommand{\fwk}[1]{f_{\mathbf{w}_{#1}}}
\newcommand{\normltwo}[1]{\left\lVert#1\right\rVert}
\newcommand{\dtwofdutwo}{\frac{\partial^{2}f}{\partial u^{2}}}
\newcommand{\dtwofdvtwo}{\frac{\partial^{2}f}{\partial v^{2}}}
\newcommand{\dtwofdudv}{\frac{\partial^{2}f}{\partial u\partial v}}
\newcommand{\curvm}{c_{\text{mean}}}
\newcommand{\curvg}{c_{\text{gauss}}}
\newcommand{\map}{\mathcal{F}}

\newcommand{\Lchd}{\mathcal{L}_{\text{CHD}}}

\newcommand{\LE}{\mathcal{L}_{E}}
\newcommand{\LG}{\mathcal{L}_{G}}
\newcommand{\Ldeform}{\mathcal{L}_{\text{def}}}
\newcommand{\Loverlap}{\mathcal{L}_{\text{ol}}}
\newcommand{\Lskew}{\mathcal{L}_{\text{sk}}}
\newcommand{\Lstretch}{\mathcal{L}_{\text{str}}}

\newcommand{\Eik}{E_{i}^{(k)}}
\newcommand{\Fik}{F_{i}^{(k)}}
\newcommand{\Gik}{G_{i}^{(k)}}
\newcommand{\Ak}{A^{(k)}}
\newcommand{\Ask}{A_{s}^{(k)}}

\newcommand{\muE}{\mu_{E}}
\newcommand{\muG}{\mu_{G}}
\newcommand{\sumKM}{\frac{1}{KM}\sum_{k=1}^{K}\sum_{i=1}^{M}}
\newcommand{\alphE}{\alpha_{E}}
\newcommand{\alphG}{\alpha_{G}}
\newcommand{\alphskew}{\alpha_{\text{sk}}}
\newcommand{\alphstr}{\alpha_{\text{str}}}
\newcommand{\alphdef}{\alpha_{\text{def}}}
\newcommand{\alpholap}{\alpha_{\text{ol}}}
\newcommand{\Lmain}{\mathcal{L}}

\newcommand{\muA}{\mu_{A}}
\newcommand{\mcol}{m_{\text{col}}}
\newcommand{\molap}{m_{\text{olap}}}
\newcommand{\mAE}{m_{ae}}
\newcommand{\mH}{m_{H}}
\newcommand{\mK}{m_{K}}
\newcommand{\DE}{D_{E}}
\newcommand{\DG}{D_{G}}
\newcommand{\Dsk}{D_{\text{sk}}}
\newcommand{\Dstr}{D_{\text{str}}}
\newcommand{\cb}{\mathbf{c}}

\newcommand{\BASIC}{\textbf{BASIC}}
\newcommand{\OURS}{\textbf{OURS}}
\newcommand{\PCAE}{\textbf{PCAE}}
\newcommand{\SC}{\textbf{SC}}
\newcommand{\SVR}{\textbf{SVR}}
\newcommand{\AN}{AN}

%% file: tex/abstract.tex
% !TEX root = ../top.tex
% !TEX spellcheck = en-US

%%%%%%%%% ABSTRACT
\begin{abstract}

Generative models that produce point clouds have emerged as a powerful tool to represent 3D surfaces, and the best current ones rely on learning an ensemble of parametric representations. Unfortunately, they offer no control over the deformations of the surface patches that form the ensemble and thus fail to prevent them from either overlapping or collapsing into single points or lines. As a consequence, computing shape properties such as surface normals and curvatures becomes difficult and unreliable.

In this paper, we show that we can exploit the inherent differentiability of deep networks to leverage differential surface properties during training so as to prevent patch collapse and strongly reduce patch overlap. Furthermore, this lets us reliably compute quantities such as surface normals and curvatures. We will demonstrate on several tasks that this yields more accurate surface reconstructions than the state-of-the-art methods in terms of normals estimation and amount of collapsed and overlapped patches.

\end{abstract}

%% file: tex/introduction.tex
% !TEX root = ../top.tex
% !TEX spellcheck = en-US

\begin{figure*}[h]
	\centering
	\includegraphics[width=0.95\linewidth]{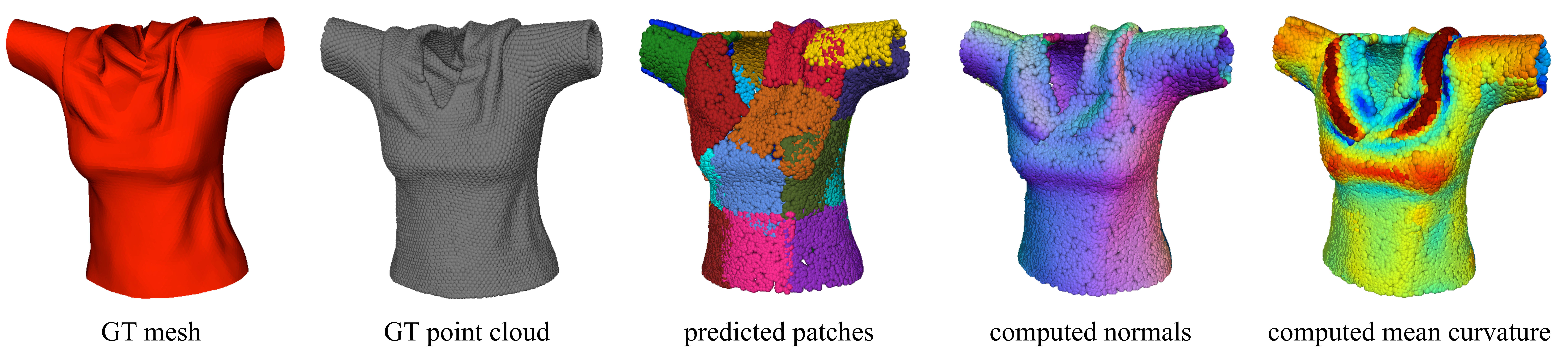}
	\caption{{\bf Our method allows for prediction of multi-patch representation where the patches are guaranteed not to collapse and to minimize the overlap.} Thanks to explicit access to the differentiable surface properties, our method computes  normals and curvature \textit{analytically} for any predicted point of the modeled surface.}
	\label{fig:teaser}
\end{figure*}

\section{Introduction}
\label{sec:introduction}

Point clouds are becoming increasingly popular as a compact and expressive way to represent 3D surfaces because they can capture high frequency geometric details without requiring much memory. State-of-the-art methods rely on encoder/decoder architectures to create latent representations from input data and then decode them using one or more learned mappings from a 2D parameter space to the 3D surface. Each one of these mappings can be thought of as transforming a 2D rectangular patch into a set of 3D points lying on the surface to be modeled. FoldingNet~\cite{Yang18a} and AtlasNet~\cite{Groueix18} are among the best representatives of this approach, and the move from one single patch to multiple ones has proved effective to achieve higher accuracy. 

However, this increase in accuracy comes at a price. Nothing guarantees that each patch will represent a substantial portion of the target surface and some may in fact collapse, meaning that they generate a single point or a line instead of a surface-like cloud. Another potential problem is that the 3D clouds generated by different patches will overlap so that the same parts of the underlying surface are represented by several patches, thus resulting in potential inconsistencies across the patches and ineffectively using the decoder's capacity. While these problems may not occur when the training data contains many diverse categories of objects, such as when using the whole of ShapeNet dataset~\cite{Chang15}, they become apparent in practical scenarios where one aims to model the shape of a specific surface, such as a piece of clothing, as shown in Fig.~\ref{fig:teaser}.

In this paper, we address these two issues by leveraging the observation that first and second derivatives of the decoder output can be used to compute the differential properties of the reconstructed surface, without having to triangulate it. In other words, we can compute \emph{exact} surface properties \emph{analytically}, rather than having to approximate these quantities using the point cloud or a mesh. This enables us to incorporate into our training loss function terms that prevent patch collapse and strongly reduce patch overlap. 

In our experiments, we will show that being able to compute differential properties and to exploit them during training (1) fully prevents any type of patch collapse, (2) substantially reduces the amount of patch overlap (3) lets us predict surface normals with higher accuracy than SotA.

Our approach to exploiting differentiability is not tied to a specific architecture and we will show on several tasks that it yields not only state-of-the-art accuracy but but also a much better behaved surface representation whose differentiable properties can be estimated easily.

Our contribution is therefore a generic approach to leveraging 3D point cloud generating schemes so that the differential properties of the target surfaces are immediately available without further post-processing, which makes them usable by subsequent algorithms that require them to perform tasks such as shape-from-shading, texture mapping, surface normal estimation  from range scans \cite{Badino11,Holzer12}, and detail-preserving re-meshing \cite{Bommes13}. 

%% file: tex/related_work.tex
% !TEX root = ../top.tex
% !TEX spellcheck = en-US

\section{Related Work}
\label{sec:related_work}

\paragraph{Deep generative approaches for surface reconstruction.}

Modern generative Deep Nets are very good at reconstructing 3D surfaces for tasks such as shape completion from incomplete data~\cite{Dai17a,Rock15,Soltani17,Shin18}, single-view shape reconstruction~\cite{Groueix18,Pumarola18,Bednarik18,Fan17a}, and auto-encoding point-clouds \cite{Groueix18,Deprelle19}. They represent the surfaces in terms of voxels \cite{Wu17,Soltani17,Firman16}, triangular meshes~\cite{Pumarola18,Bednarik18,Danerek17}, or point clouds~\cite{Groueix18,Deprelle19,Fan17a}. The common denominator of all these methods is that they deliver precise shapes in terms of 3D locations but not necessarily in terms of differential surface properties, such as normals and curvature. The latter may be inaccurate and even nonsensical as will be shown in the experiment section. 

\parag{Patch-based representations.}

Among all these methods, FoldingNet~\cite{Yang18a} was the first to introduce the idea of learning a parametric mapping from a 2D patch to a 3D surface. It relies on a discrete sampling and follow-up methods introduced continuous multi-patch representations  that are trained by minimizing  the Chamfer distance \cite{Groueix18,Deprelle19}, a shape aware variant of the L2 distance \cite{Sinha17},  or are optimized to predict a single sample using Earth mover's distance\cite{Williams19}. One of the biggest advantage of these approaches is that the learned mapping, being a continuous function, 
allows for arbitrary sampling resolution at test time. However, still none of these methods gives 
access to the differential surface properties. An exception is the approach of~\cite{Laube18} 
that learns a parameterization for B-spline approximation but only works with 2D curves.

\parag{Using differential surface properties for training.}

There are a few deep learning techniques that use differential surface properties in the form of either normal maps \cite{Bednarik18,Bansal16} or their approximations computed on triangular meshes~\cite{Gundogdu19} but none that rely on 3D point clouds. Using differential surface properties still mostly belongs to the realm of non-deep learning methods, for example for shape from template~\cite{Ngo16,Bartoli15} or non-rigid structure-from-motion \cite{Agudo16,Parashar19a}, which are beyond the scope of this paper. 

%% file: tex/multi_patch_representations.tex
% !TEX root = ../top.tex
% !TEX spellcheck = en-US

%%%%%%%%%%%%%%%%%%%%%%%%%%%%%%%%%%%%%%%%%%%%%%%%%%%%%%%%%%%%%%%%%%%%%%%%%%%%%%%%%%%%%%%%%%%%%%%%%%%%
\section{Multi Patch Representations}

As discussed above, multi-patch representations~\cite{Groueix18,Williams19} are powerful tools for 
generating surfaces from latent representations. However, they suffer from a number of limitations 
that we discuss below and will address in Section~\ref{sec:methodology}.  

%%%%%%%%%%%%%%%%%%%%%%%%%%%%%%%%%%%%%%%%%%%%%%%%%%%%%%%%%%%%%%%%%%%%%%%%%%%%%%%%%%%%%%%%%%%%%%%%%%%%
\subsection{Formalization}
\label{sec:formalization}

Let us consider a mapping $\map$ from a given low-dimensional latent vector  $\db \in \real^{D}$ 
to a surface $\surf$ in 3D space represented by a cloud of 3D points. In the multi-patch approach, 
the point cloud is taken to be the union of points generated by $K$ independent mappings 
$\fwk{k}:~\real^D \times \domf \rightarrow~\real^{3}$ for $ 1 \leq j \leq K$, where each $\fwk{k}$ 
is a trainable network with parameters $\wb_{k}$ and $\domf = [c_{\text{min}},c_{\text{max}}]^{2}$ 
represents a square in $\real^{2}$. In other words, the $\fwk{k}$ network takes as input a latent 
vector $\db$ and a 2D point in $\domf$ and returns a 3D point. 

Given a training set containing many 3D shapes, the  $\wb_{k}$ network weights are learned by minimizing a sum of chamfer-based losses, one for each 3D shape in a training batch, of the form
\begin{small}
\begin{equation}
\begin{split}
\Lchd = &\frac{1}{KM}\sum_{k=1}^{K}\sum_{i=1}^{M}\min_{j} \normltwo{\pbi^{(k)} - \qbj}^{2} + \\ 
&\frac{1}{N}\sum_{j=1}^{N}\min_{i,k}\normltwo{\pbi^{(k)} - \qbj}^{2},
\end{split}
\label{eq:chamfer_distance}
\end{equation}
\end{small}
where $M$ is the number of points predicted by each patch, $N$ is the number of GT points, $\pbi^{k}$ is the $i$-th 3D point predicted by $\fwk{k}$, and $\qbj$ is the $j$-th GT point. The whole pipeline is depicted in Figure \ref{fig:pipeline}.

%%%%%%%%%%%%%%%%%%%%%%%%%%%%%%%%%%%%%%%%%%%%%%%%%%%%%%%%%%%%%%%%%%%%%%%%%%%%%%%%%%%%%%%%%%%%%%%%%%%%
\subsection{Limitations}
\label{sec:limitations}

Minimizing the loss function of Eq.~\ref{eq:chamfer_distance} yields patches 
that jointly cover the whole surface but does not constrain how much deformation individual patches 
undergo or how they are positioned with respect to each other. In practice, this leads to two 
potential failure modes. 

\input{fig/collapse}

\paragraph{Patch Collapse.} 

Some of the patches might collapse to undesirable configurations, such as those shown in Figure~\ref{fig:collapse}. This may not increase the total $\Lchd$ value much when training is complete because the remaining, non-collapsed patches can compensate for the collapsed ones. However, collapsed patches still cause two main problems. First, their 
normals and curvature do not make sense anymore. Second,  the $\fwk{j}$ corresponding to a collapsed patch becomes useless, thus wasting the representational power of $\map$. 
\paragraph{Patch Overlap.} 

\input{fig/overlap}

Another drawback of this approach is that it provides no control over the relative spatial configuration of the patches. Thus, it does not prevent them from overlapping each other, as depicted by Figure \ref{fig:overlap}. This is an  undesirable behavior because the overlapping patches may yield surfaces that are not well aligned and, again, because some of the expressive power of $\map$ is wasted.

%% file: fig/collapse.tex
% !TEX root = ../top.tex
% !TEX spellcheck = en-US

\begin{figure}[H]
	\centering
	\includegraphics[width=0.99\linewidth]{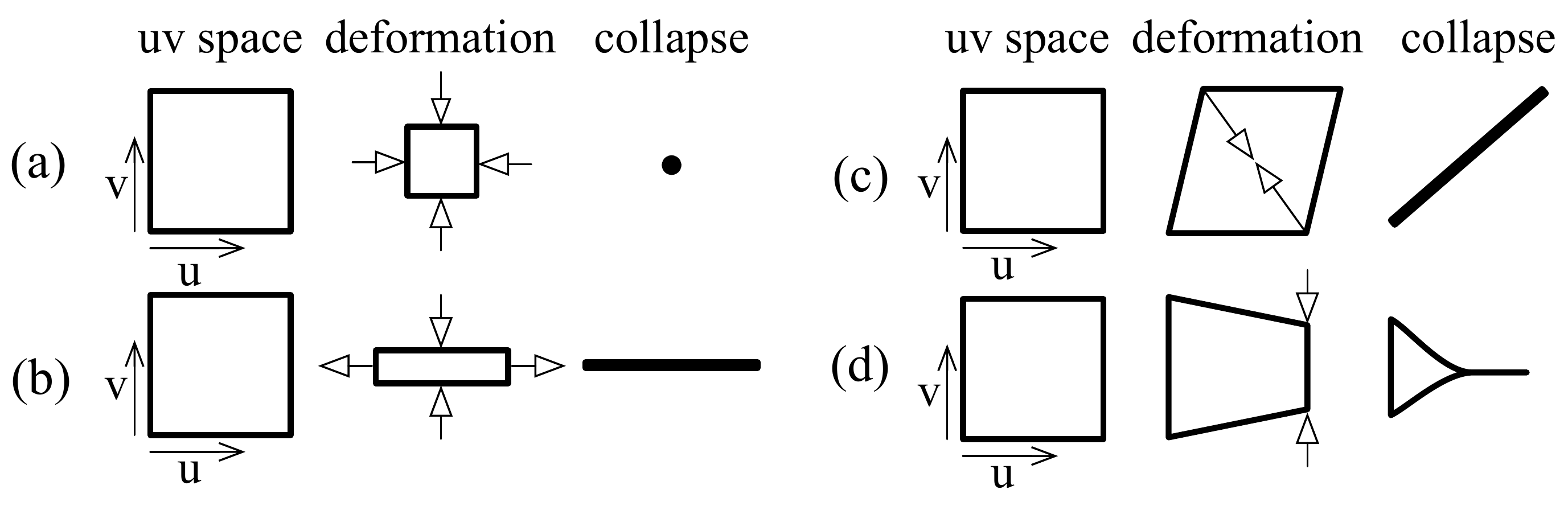}
	\caption{{\bf 2D representation of typical patch collapses.} (a) 0D collapse, (b) 1D stretch collapse, 
		(c) 1D skew collapse and (d) partial collapse.}
	\label{fig:collapse}
\end{figure}

%% file: fig/overlap.tex
% !TEX root = ../top.tex
% !TEX spellcheck = en-US

\begin{figure}[h]
	\centering
	\includegraphics[width=0.9\linewidth]{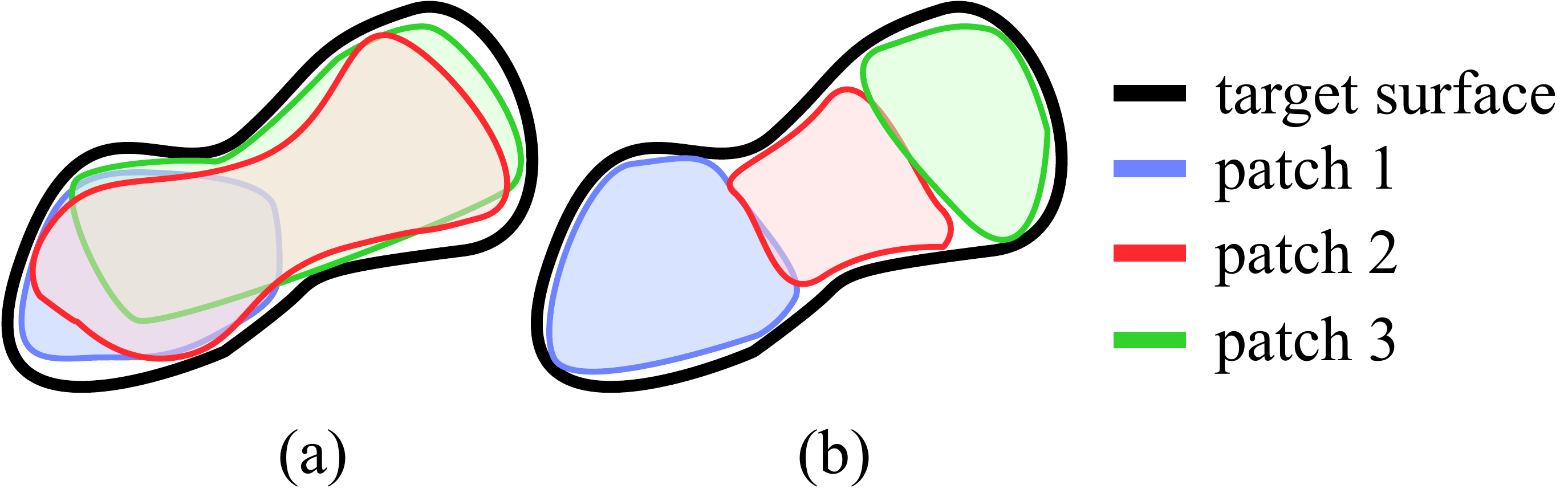}
	\caption{{\bf Patch overlap}. (a) Up to 3-fold overlap. (b) Configuration with small overlap.}
	\label{fig:overlap}
\end{figure}

%% file: tex/differnetiable_properties.tex
% !TEX root = ../top.tex
% !TEX spellcheck = en-US

\input{fig/pipeline}

%%%%%%%%%%%%%%%%%%%%%%%%%%%%%%%%%%%%%%%%%%%%%%%%%%%%%%%%%%%%%%%%%%%%%%%%%%%%%%%%%%%%%%%%%%%%%%%%%%%%
\section{Accounting for Differential Properties} 
\label{sec:methodology}

We have seen that multi-patch representations are powerful but are subject to patch collapse and 
overlap, both of which reduce their expressive power. We now show that by regularizing the 
differential properties of the reconstructed surfaces during training, we can eliminate patch 
collapse and mitigate patch overlap. We will demonstrate in Section~\ref{sec:experiments} that this boosts the accuracy of
normals and curvature.

In the remainder of this section, we first explain how we can compute online the differential properties of surfaces represented by a 3D point cloud. We then present our approach to using them during training.

%%%%%%%%%%%%%%%%%%%%%%%%%%%%%%%%%%%%%%%%%%%%%%%%%%%%%%%%%%%%%%%%%%%%%%%%%%%%%%%%%%%%%%%%%%%%%%%%%%%%
\subsection{Differential Surface Properties} 
\label{sec:differential_surface_properties}

Let $\rb = [u, v] \in \domf$, where $\domf$ is the 2D domain over which $\fwk{k}$ is defined, as explained at the beginning of Section~\ref{sec:formalization}. $\pb=\fwk{k}(\rb)$ is a point of surface $\surf$. We can compute the differential properties of $\surf$, including normals and curvatures,  from the derivatives of $\fwk{k}$ with respect to $u$ and $v$ as follows, given that $\fwk{k}$ is a continuously differentiable function.
For notational simplicity, we drop the subscript $\wb_{k}$ from $\fwk{k}$ in the remainder of this section.

Let $\jac = \begin{bmatrix} \fu & \fv \end{bmatrix}$  be  the Jacobian of $f$ at $\pb$, where $\fu = \frac{\partial f}{\partial u}$ and $\fv = \frac{\partial f}{\partial v}$. The normal vector is %
\begin{align}
\label{eq:normals}
\nb = \frac{\fu \times \fv}{\normltwo{\fu \times \fv }}.
\end{align} 
The curvature, area and deformation properties can be computed from the \textit{metric tensor} 
\begin{align}
g = \jac^{\top}\jac = 
	\begin{bmatrix} \fu^{\top}\fu & \fu^{\top}\fv \\ 
					\fu^{\top}\fv & \fv^{\top}\fv 
	\end{bmatrix} = 
	\begin{bmatrix} E & F \\ 
					F & G 
	\end{bmatrix}.
\label{eq:metricTensor}
\end{align}
The mean and Gaussian curvature are then 
\begin{align}
\label{eq:curvature_mean}
\curvm &= -\frac{1}{2 \det{g}}\nb^{\top} 
\left[ \dtwofdutwo G - 2\dtwofdudv F + \dtwofdvtwo E \right], \\
\label{eq:curvature_gauss}
\curvg &= \frac{\dtwofdutwo^{\top} \nb \cdot 
	\dtwofdvtwo^{\top} \nb - (\dtwofdudv^{\top}\nb)^{2}}{EG - F^{2}}.
\end{align}
Furthermore, the area of the surface covered by the patch can be estimated as

\begin{align}
A = \iint_{\domf}{\sqrt{EG - F^{2}}dudv}.
\label{eq:patchArea}
\end{align}
Note that all these surface properties are computed \textit{analytically}. Thus they are \textit{exact}, differentiable, and do not require a triangulation. This is unlike traditional methods that \textit{approximate} these quantities on a point cloud or a mesh.

%%%%%%%%%%%%%%%%%%%%%%%%%%%%%%%%%%%%%%%%%%%%%%%%%%%%%%%%%%%%%%%%%%%%%%%%%%%%%%%%%%%%%%%%%%%%%%%%%%%%
\subsection{Learning a Robust Mapping}

Recall that our goal is to learn a mapping $\map$ from multiple 2D patches to a 3D point cloud. In particular, we seek to constrain the deformations modeled by $\map$ such that patch collapse is
prevented but complex surfaces can still be represented accurately. We now discuss the deformation model that we rely on and then introduce the required training loss functions.

\subsubsection{Deformation Model} 
\label{sec:deformations}

Conformal mappings yield low distortions while retaining the capacity to model complex shapes. They are therefore widely used in computer graphics and computer vision, for example for texture mapping~\cite{Botsch10} and surface reconstruction~\cite{Parashar19a}. For a surface to undergo conformal deformation, the metric tensor must be of the form
\begin{equation}
g_{\text{conf}} = s(\rb) \begin{bmatrix} 1 & 0 \\ 0 & 1 \end{bmatrix},
\end{equation}
where $s: \domf \rightarrow \real$ returns a scale value for each position in the parameter space. Unfortunately, making the deformation conformal does not prevent patch collapse, even in a single-patch scenario, since partial collapse can still occur wherever $s(\rbi) \sim 0$.

To address this, we propose to use \textit{fixed-scale conformal} mappings whose metric tensors can be written as
\begin{equation}
g_{\text{fsconf}} = s \begin{bmatrix} 1 & 0 \\ 0 & 1 \end{bmatrix},
\label{eq:scaledMetricTensor}
\end{equation}
where $s$ is an unknown global scale shared by all parameter space locations. Constraining $\map$ to be a \textit{fixed scale conformal} mapping amounts to assuming that the
target surface is patch-wise developable, which has proved to be a reasonable assumption in the domain 
of deformable surface reconstruction~\cite{Bartoli15,Bartoli12b,Ngo15b}.

In a single patch scenario, $s \sim 0$ is not an option anymore when minimizing the loss of Eq.~\ref{eq:chamfer_distance} because the resulting surface would be point-collapsed and thus could not cover the full target surface. However, collapses can still occur in the multi-patch case, and have to be prevented using appropriate loss terms, as discussed below.

%%%%%%%%%%%%%%%%%%%%%%%%%%%%%%%%%%%%%%%%%%%%%%%%%%%%%%%%%%%%%%%%%%%%%%%%%%%%%%%%%%%%%%%%%%%%%%%%%%%%
\subsubsection{Loss Functions} 
\label{sec:loss_functions}

Here, we formulate the loss terms that should be added to the data loss $\Lchd$ of Eq.~\ref{eq:chamfer_distance} during training to ensure that the resulting $\fwk{k}$ are fixed-scale conformal, as described above, without patch collapse, and with as little overlap as possible. 

\parag{Enforcing Conformality.} 

We define
\begin{align}
	\LE &= \sumKM\left(\frac{\Eik - \muE}{\Ak}\right)^{2}, \\
	\LG &= \sumKM\left(\frac{\Gik - \muG}{\Ak}\right)^{2}, \\
	\Lskew &= \sumKM\left(\frac{\Fik}{\Ak}\right)^{2}, \\
	\Lstretch &= \sumKM\left(\frac{\Eik - \Gik}{\Ak}\right)^{2},
\end{align}
where $M$ is the number of points sampled on each surface patch; $E$, $F$, and $G$ are defined in Eq.~\ref{eq:metricTensor}; $\muE~=~\frac{1}{KM}\sum_{k}\sum_{i}\Eik$ and $\muG~=~\sum_{k}\sum_{i}\Gik$; and $\Ak$ is the area of a surface patch computed using Eq.~\ref{eq:patchArea}. Note that we normalize these terms by $\Ak$ to make them independent of the current surface patch area, which changes over the course of training.

Each one of these four losses controls the components of the metric tensor  of Eq.~\ref{eq:metricTensor}, so that its off-diagonal terms are close to $0$ and its diagonal terms are equal as in Eq~\ref{eq:scaledMetricTensor}. Concretely, $\Lstretch$ prevents the 0D and 1D collapses shown in Fig.~\ref{fig:collapse}(a,b).
$\Lskew$ prevents 1D skew collapse as depicted by Fig.~\ref{fig:collapse}(c) while $\LE$ and $\LG$ prevent partial ones depicted by Fig.~\ref{fig:collapse}(d). Finally, we express our complete deformation loss as
\begin{equation}
\Ldeform = \alphE\LE + \alphG\LG + \alphskew\Lskew + \alphstr\Lstretch \; , \label{eq:defLoss}
\end{equation} 
where $\alphE, \alphG, \alphskew, \alpholap \in \real$ are hyperparameters.

\parag{Minimizing Overlaps.} 

Recall from Section~\ref{sec:differential_surface_properties}, that we can compute the area $\Ak$ covered by a patch $k$ using Eq.~\ref{eq:patchArea}. We therefore introduce
\begin{align}
	\Loverlap = \max\left(0, \sum_{k=1}^{K}\left(\Ak\right) - \hat{A}\right)^{2} 
\end{align}
to encourage the patches to jointly cover at most the area of the entire surface, where $\hat{A}$ is computed as the area of the GT mesh or of a triangulated GT depth map, depending on the task of interest.
We estimate the patch area as $\Ak = \frac{1}{M^{(k)}}\sum_{i=1}^{M^{(k)}} \Ak_{i}$, where $M^{(k)}$ is
the number of points sampled from the patch $k$ and $A^{(k)}_{i}$ is Eq.~\ref{eq:patchArea} computed for a single point.

\parag{Combined Loss Function.} 

We take our complete loss function to be
\begin{equation}
\Lmain = \Lchd + \alphdef\Ldeform + \alpholap\Loverlap \; , \label{eq:mainLoss}
\end{equation}
with hyperparameters $\alphdef, \alpholap \in \real$. In practice, we use the weights of Eq.~\ref{eq:defLoss} to control the relative influence of the four terms of $\Ldeform$, and $\alphdef$ and $\alpholap$ to control the overall magnitude of the deformation and overlap regularization term respectively.

%%%%%%%%%%%%%%%%%%%%%%%%%%%%%%%%%%%%%%%%%%%%%%%%%%%%%%%%%%%%%%%%%%%%%%%%%%%%%%%%%%%%%%%%%%%%%%%%%%%%
\subsubsection{Mapping Architecture} 
\label{sec:parametric_mapping_formulation}

\begin{figure}[htb]
	\centering
	\includegraphics[width=0.99\linewidth]{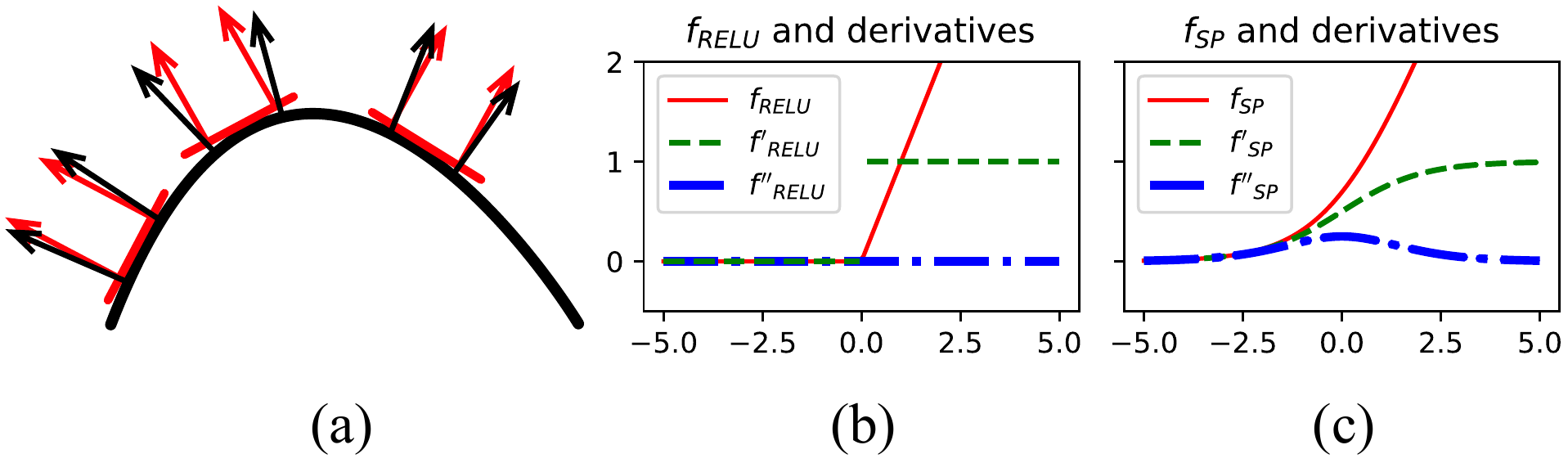}
	\caption{The use of the ReLU results in a piecewise linear mapping, which, as shown in red in (a), only poorly approximates the true surface normals (in black). As can be seen by comparing (b) and (c), the Softplus function approximates the ReLU behavior, while having smooth 1st and 2nd derivatives.}
	\label{fig:relu_sp}
\end{figure}

As in~\cite{Groueix18}, we implement each mapping $\fwk{k}$ as a multi-layer perceptron (MLP), with each MLP having its own set of weights. As we need $\fwk{k}$ to be  at least $C^{2}$-differentiable, we cannot use the popular ReLU activation function, which is $C^{2}$ only \emph{almost} everywhere. Note that the ReLU function would also yield a piecewise linear mapping, which would be ill-suited to compute surface curvatures. Therefore, we use the \textit{Softplus} function, which approximates the ReLU while having smooth 1st and the 2nd order derivatives, as shown in Figure~\ref{fig:relu_sp}.

%% file: fig/pipeline.tex
% !TEX root = ../top.tex
% !TEX spellcheck = en-US

\begin{figure*}[htb]
	\centering
	\includegraphics[width=0.99\linewidth]{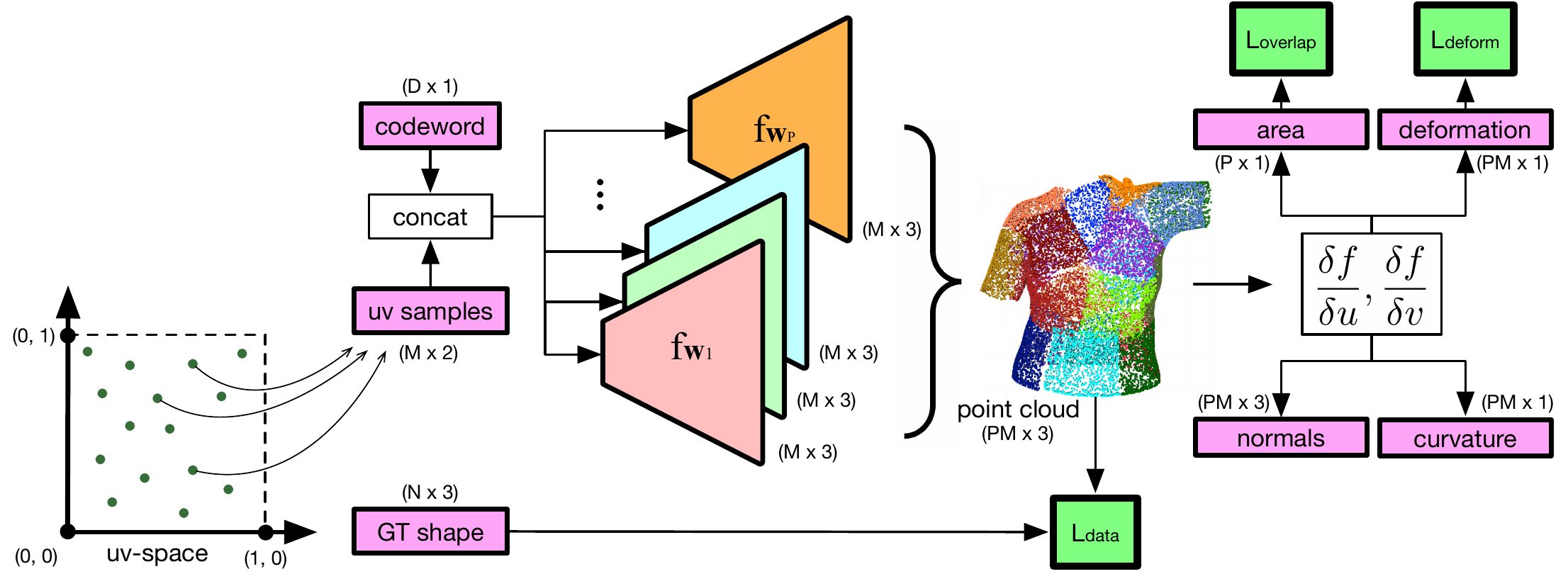}
	\caption{{\bf Schematic view of our approach.} The input to the set of $K$ decoders $\{\fwk{k}\}$ is a 
		latent vector $\db$ (dubbed \textit{codeword}) and a set of 2D points 
		sampled from the domain ${\cal D}_f$ (dubbed \textit{uv-space}). The decoders produce point
	 	clouds, which together represent the target surface. The derivatives of $\fwk{k}$ w.r.t. the 
	 	\textit{uv-space} allow for the analytical computation of each patch's area and deformation 
	 	properties. The loss function used to train our model consists of a data term $\mathcal{L}_{\text{data}}$ using GT 
	 	annotations and of terms $\mathcal{L}_{\text{ol}}$ and $\mathcal{L}_{\text{def}}$ 
	 	which prevent patch overlap and patch
 	    collapse, respectively.}
	\label{fig:pipeline}
\end{figure*}

%% file: tex/experiments.tex
% !TEX root = ../top.tex
% !TEX spellcheck = en-US

%%%%%%%%%%%%%%%%%%%%%%%%%%%%%%%%%%%%%%%%%%%%%%%%%%%%%%%%%%%%%%%%%%%%%%%%%%%%%%%%%%%%%%%%%%%%%%%%%%%%
\section{Experiments}
\label{sec:experiments}

Our approach is generic and can thus be applied to different tasks relying on different architectures.
We discuss them below, and then introduce the datasets and metrics we use for testing purposes. Finally, we present our results. 

%%%%%%%%%%%%%%%%%%%%%%%%%%%%%%%%%%%%%%%%%%%%%%%%%%%%%%%%%%%%%%%%%%%%%%%%%%%%%%%%%%%%%%%%%%%%%%%%%%%%
\subsection{Tasks}
\label{sec:tasks}

We experimented with three popular surface reconstruction tasks, which we describe below, together with the architectures we used to tackle them. 

\parag{Point Cloud Autoencoding (\PCAE{}).} 

We rely on an AtlasNet~\cite{Groueix18} variant in which we slightly modified the decoder: The ReLU activations are replaced with Softplus for the reasons stated in Section~\ref{sec:parametric_mapping_formulation} and the last activation is linear. We removed the batch normalization layers because they made our training unstable.

\parag{Shape completion (\SC{}).} 

Given a partial 3D point cloud, such as a depth map, shape completion aims to predict a complete object shape. To this end, we use a U-Net~\cite{Ronneberger15} encoder that produces a latent representation $\db$ of size $2048$ and the FoldingNet~\cite{Yang18a} decoder with all the activations replaced with Softplus except the last one which is linear.

\parag{Single-View 3D Shape Reconstruction (\SVR{}).}  

The goal of this task is to predict the shape of a surface observed in a single RGB image. To this end, we use an encoder implemented as a ResNet~\cite{He16} with bottleneck modules and $44$ layers and a decoder implemented as a FoldingNet~\cite{Yang18a} variant similar to the one used for \SC{}. For the sake of fair comparison with the SotA, we also experimented with a variant of AtlasNet \cite{Groueix18} with the same modifications as in \PCAE{}. 
\\~\\~
In the remainder of this section, regardless of the task at hand, we will refer to our model which we train using
$\Lmain$ of Eq.~\ref{eq:mainLoss} as \OURS{}. For every experiment we report what values we used for the
hyperparameters $\alphdef$ and $\alpholap$ of Eq.~\ref{eq:mainLoss}. Unless stated otherwise, we
set all the hyperparameters of Eq~\ref{eq:defLoss} to be equal to $1$.

%%%%%%%%%%%%%%%%%%%%%%%%%%%%%%%%%%%%%%%%%%%%%%%%%%%%%%%%%%%%%%%%%%%%%%%%%%%%%%%%%%%%%%%%%%%%%%%%%%%%
\subsection{Datasets}
\label{sec:datasets}

%We use the following three datasets.
\vspace{0.2cm}
\parag{ShapeNet Core v2~\cite{Chang15} (SN).} This dataset consists of synthetic objects of multiple 
categories and has been widely used to gauge the performance of 3D shape reconstruction approaches. 
We use the same train/test split as in the AtlasNet paper~\cite{Groueix18}.

\parag{Textureless deformable surfaces~\cite{Bednarik18} (TDS).} This real-world dataset of deformable surfaces captured in various lighting conditions consists of sequences of RGB images and  corresponding depth and normal maps of $5$ different objects. We selected the $2$ for which the most data samples are available, a piece of cloth and a T-Shirt imaged with a single light setup. We use  $85\%$ of the samples for training, $5\%$ for validation, and $10\%$ for testing. 

\parag{Female T-Shirts \cite{Gundogdu19} (FTS).} This synthetic dataset comprises T-Shirts worn by $600$ different women in different poses. We randomly split the body shapes into 
training ($87\%$), validation ($8\%$) and testing ($5\%$) sets. We precomputed the mean and Gaussian curvature on the GT meshes using quadric fitting implemented in the graphics library libigl \cite{Jacobson18}.

%%%%%%%%%%%%%%%%%%%%%%%%%%%%%%%%%%%%%%%%%%%%%%%%%%%%%%%%%%%%%%%%%%%%%%%%%%%%%%%%%%%%%%%%%%%%%%%%%%%%
\subsection{Metrics}
\label{sec:metrics}

We report our results in terms of the following metrics. 

\parag{Chamfer Distance (CHD).} This distance is given in Eq.~\ref{eq:chamfer_distance}.

\parag{Mean ($\mathbf{\mH}$) and Gaussian ($\mathbf{\mK}$) Curvature.} The curvatures are given in Eq.~\ref{eq:curvature_mean} and \ref{eq:curvature_gauss}.

\parag{Angular error ($\mathbf{\mAE}$).} To measure the accuracy of the computed normals, we define
the mean angular error as $\mAE = \frac{1}{M}\sum_{i=1}^{M} \arccos{|\nb_{i}\hat{\nb_{i}}|}$, where 
$\nb_{i}$ and $\hat{\nb_{i}}$ are the unit length normals of a predicted point and its closest GT point.
The absolute value is taken to make the metric invariant to the patch orientation, which neither our method
nor the SotA approaches controls.

\parag{Number of collapsed patches ($\mathbf{\mcol}$)} We assume a patch $k$ to be collapsed if $\Ak < c_{A}\muA$, where
$\muA = \sum_{k=1}^{K}\Ak$ is the mean patch area and $c_{A}$ is a constant to be chosen. We define the
patch collapse metric as $\mcol = \frac{1}{S}\sum_{s=1}^{S}\sum_{k=1}^{K}\mathbb{I}_{\left[ \Ask < c_{A}\muA \right]}$,
that is, an average number of collapsed patches over a dataset of size $S$. In all our experiments, we 
set $c_{A} = 1e^{-3}$.

\parag{Amount of overlap ($\mathbf{\molap^{(t)}}$).} For each ground-truth 3D point, 
we count the number of patches within a predefined threshold $t$ and take the average over all the points.

%%%%%%%%%%%%%%%%%%%%%%%%%%%%%%%%%%%%%%%%%%%%%%%%%%%%%%%%%%%%%%%%%%%%%%%%%%%%%%%%%%%%%%%%%%%%%%%%%%%%
\subsection{Normal and Curvature Estimates}
\label{sec:normals_and_curvature}

As discussed in \ref{sec:limitations},  collapsed patches can strongly affect the quality of the normals and curvatures we can recover from estimated surfaces. \OURS{} for $\SC$ on FTS prevents the collapses from happening. To demonstrate this, we trained the network also without the deformation loss $\Ldeform$ term of Eq.~\ref{eq:defLoss}, to which we refer as \BASIC{}. We used the FTS dataset for all our experiments. For training we randomly sampled $8000$ GT points and the same number is predicted by \OURS{}.

\begin{figure}[t]
	\centering
	\includegraphics[width=0.85\linewidth]{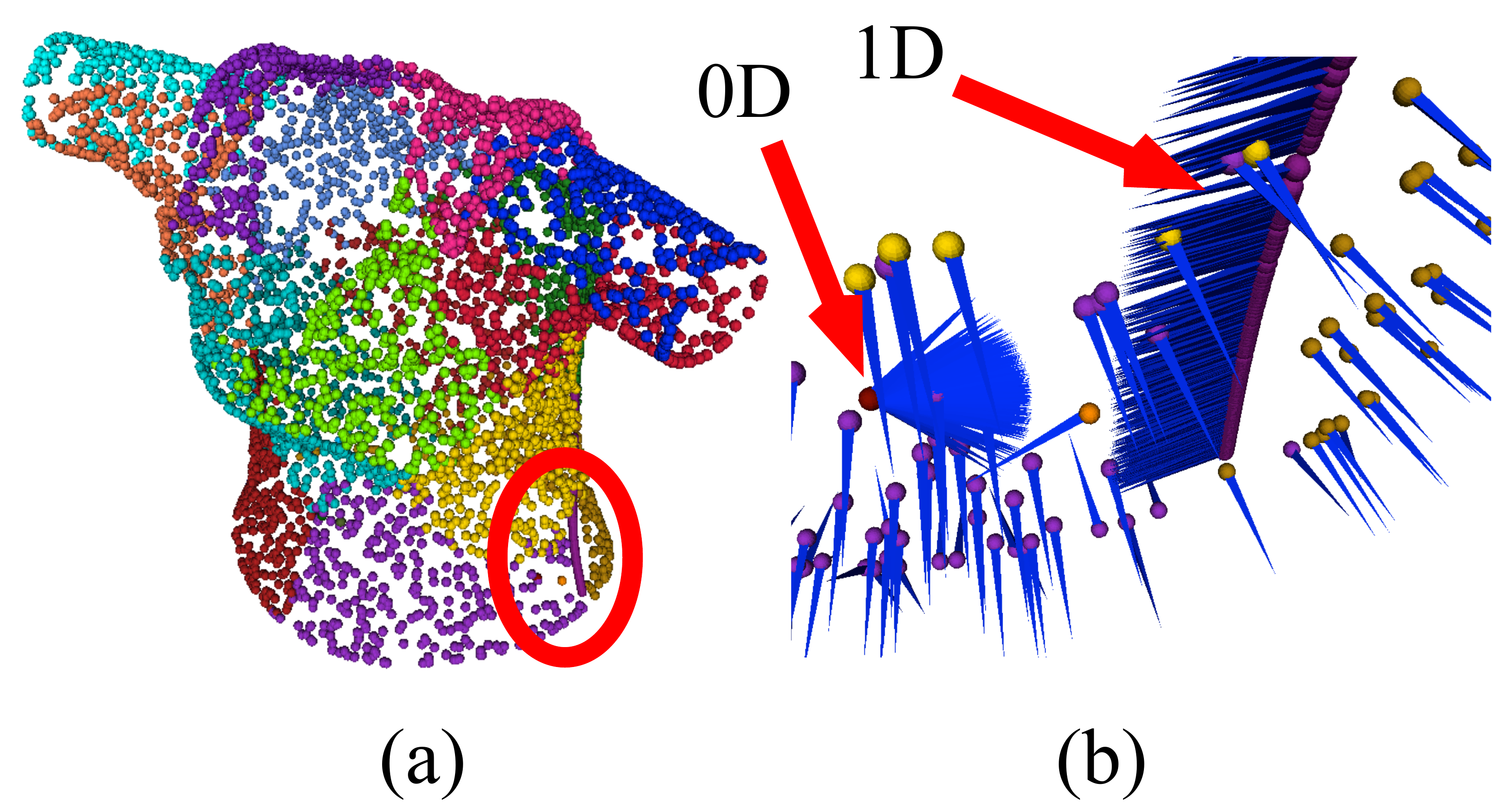}
	\caption{{\bf Typical cases of 0D and 1D collapses.} (a) Predicted point cloud with different colors 
		denoting the individual patches. The red ellipse focuses on a collapsed region. (b) Close up view of 
		the 0D and 1D collapses and corresponding normals.}
	\label{fig:collapse_point_line}
\end{figure}

We set $\alphdef = 1e^{-3}$ and $\alpholap = 0$, thus ignoring overlaps, in Eq.~\ref{eq:mainLoss} and the number of patches to $25$.
We report the results in Table~\ref{tab:normals_curvature} and Fig.~\ref{fig:collapse_point_line} depicts typical collapse cases. Because there are no collapses for \OURS{}, the resulting accuracy is improved, which validates our hypothesis that allowing patches to collapse wastes some of the networks descriptive power. Furthermore, the quality improvement of the computed normals is all the more significant. By contrast, {\it not} using the $\Ldeform$ makes the normals and curvatures, computed using Eq. \ref{eq:normals}, \ref{eq:curvature_mean} and \ref{eq:curvature_gauss}, useless, as illustrated by Fig. \ref{fig:collapse_point_line}.

\begin{table}[t]
  \centering
  \caption{{\bf Training with (\OURS) and without $\Ldeform$ (\BASIC)} for \SC{} on the FTS dataset. The CHD is multiplied by $1e3$ and the $\mAE$ is expressed in degrees. Note that computing the curvatures on the surface obtained with the model trained without $\Ldeform$ suffers from  numerical instabilities, which prevented us from reporting $\mH$ value when not using $\Ldeform$.}
    \begin{tabular}{lccccc}
    \textbf{Model} & \multicolumn{1}{c}{\textbf{CHD}} & \multicolumn{1}{c}{\textbf{$\mathbf{\mAE}$}} & \multicolumn{1}{c}{\textbf{$\mathbf{\mH}$}} & \multicolumn{1}{c}{\textbf{$\mathbf{\mK}$}} & \multicolumn{1}{c}{$\mathbf{\mcol}$} \\
    \midrule
    \BASIC{} 				  & 0.14 & 24.38 & n/a   & 170e6 & 9  \\
    \OURS{}     & \textbf{0.11} & \textbf{5.94}  & \textbf{35.29} & \textbf{53e3} & \textbf{0}  \\
    \bottomrule
    \end{tabular}
  \label{tab:normals_curvature}
\end{table}

%%%%%%%%%%%%%%%%%%%%%%%%%%%%%%%%%%%%%%%%%%%%%%%%%%%%%%%%%%%%%%%%%%%%%%%%%%%%%%%%%%%%%%%%%%%%%%%%%%%%
\subsection{Number of Patches and Collapse}
\label{sec:number_of_patches_and_collapse}

It could be argued that the patch collapse problems described above when not enforcing the regularization constraints are simply a consequence of using too many patches and that using fewer would cure the problem. We now show this not to be the case. 

We trained \OURS{} for $\SVR$ and, as before, we also trained \BASIC{}, that is, no $\Ldeform$ term. We used the TDS dataset for all our experiments. For training,  we sampled $3000$ points randomly from the GT depth map and the same number is predicted by our model. As before, we set  $\alphdef = 1e^{-3}$ and $\alpholap = 0$.  In Table~\ref{tab:collapses_tds} we show that regardless of the number of patches, the collapses always occur when using \BASIC{} and never when using \OURS{}. While CHD is comparable for both models, the angular error is always lower for \OURS{}.

\begin{table}[htbp]
  \centering
  \caption{{\bf Training the model with (\OURS{}) and without $\Ldeform$ (\BASIC{})} for \SVR{} on the TDS dataset. Note that the normals are much more accurate when using  $\Ldeform$.}
	\resizebox{0.45\textwidth}{!}{
  	\begingroup
  	\setlength{\tabcolsep}{1pt}
  	\renewcommand{\arraystretch}{0.6}
\begin{tabular}{l|cccc|cccc}
	\multicolumn{1}{r}{} &   \multicolumn{4}{c}{\textbf{Cloth}} & \multicolumn{4}{c}{\textbf{T-Shirt}} \\
	\multicolumn{1}{c|}{\textbf{Model}} & \textbf{\# patch.} & \textbf{CHD} & \textbf{$\mathbf{\mAE}$} & \textbf{$\mathbf{\mcol}$} & \textbf{\# patch.} & \textbf{CHD} & $\mathbf{\mAE}$ & \textbf{$\mathbf{\mcol}$} \\
	\midrule
	\BASIC{} & 2     & 0.36  & 30.32 & 1     & 2     & \textbf{0.48} & 48.32 & 1 \\
	\OURS{} & 2     & \textbf{0.33} & \textbf{20.90} & \textbf{0} & 2     & 0.51  & \textbf{20.35} & \textbf{0} \\
	\midrule
	\BASIC{} & 3     & 0.35  & 21.95 & 1     & 3     & 0.46  & 36.29 & 1 \\
	\OURS{} & 3     & \textbf{0.32} & \textbf{20.60} & \textbf{0} & 3     & \textbf{0.49} & \textbf{20.44} & \textbf{0} \\
	\midrule
	\BASIC{} & 4     & \textbf{0.37} & 28.77 & 2     & 4     & \textbf{0.41} & 22.95 & 1 \\
	\OURS{} & 4     & 0.39  & \textbf{21.08} & \textbf{0} & 4     & 0.42  & \textbf{20.77} & \textbf{0} \\
	\midrule
	\BASIC{} & 10    & \textbf{0.33} & 25.67 & 2.02  & 15    & \textbf{0.41} & 23.51 & 2 \\
	\OURS{} & 10    & 0.41  & \textbf{20.17} & \textbf{0} & 15    & \textbf{0.41} & \textbf{20.93} & \textbf{0} \\
	\bottomrule
\end{tabular}%
	\endgroup
	}	
  \label{tab:collapses_tds}
\end{table}

%%%%%%%%%%%%%%%%%%%%%%%%%%%%%%%%%%%%%%%%%%%%%%%%%%%%%%%%%%%%%%%%%%%%%%%%%%%%%%%%%%%%%%%%%%%%%%%%%%%%
\subsection{Comparison to the SotA on PCAE and SVR}
\label{sec:comparison_to_sota}

Here we compare the predictions delivered by \OURS{} against those delivered by AtlasNet~\cite{Groueix18} (\AN) on two tasks, \PCAE{} on the ShapeNet dataset and \SVR{} on the TDS dataset. In both cases, our goal is not only to minimize CHD but also to minimize patch overlap. We again set $\alphdef = 1e^{-3}$ but now turn on the $\Loverlap$ loss by setting $\alpholap=1e^{2}$. 

\parag{Autoencoding on ShapeNet.} 

We retrained the original \AN{} using the code provided by the authors and trained \OURS{} on $\PCAE$ using the ShapeNet dataset separately on object categories airplane, chair, car, couch and cellphone. We used $25$ patches, $2500$ points randomly sampled from the GT point clouds and the same amount is predicted by \OURS{} and \AN{}. Since the ShapeNet objects often contain long thin parts (e.g. legs of a chair or wings of an airplane) we chose to allow patch stretching and set $\alphstr$ of Eq.~\ref{eq:defLoss} to $0$. We trained both \OURS{} and \AN{} until convergence.

\begin{figure}[htb]
	\centering
	\includegraphics[width=0.95\linewidth]{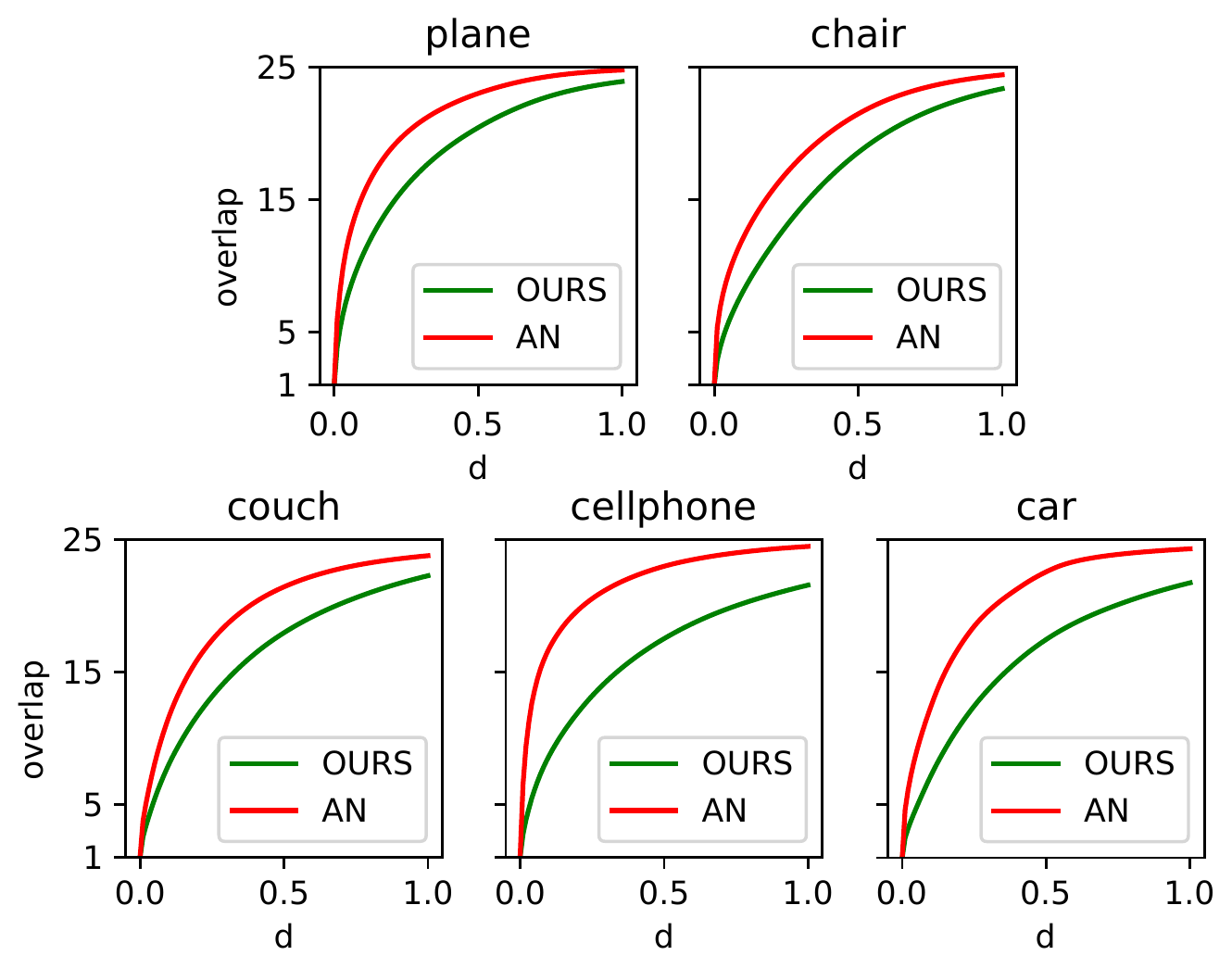}
	\caption{{\bf Patch overlap for \OURS{} and \AN{}} trained for \PCAE{} on the ShapeNet dataset. We plot $\mcol^{(t)}$ as a function of $t$.}
	\label{fig:growing_k_sn}
\end{figure}

\begin{figure}[htb]
	\centering
	\includegraphics[width=0.99\linewidth]{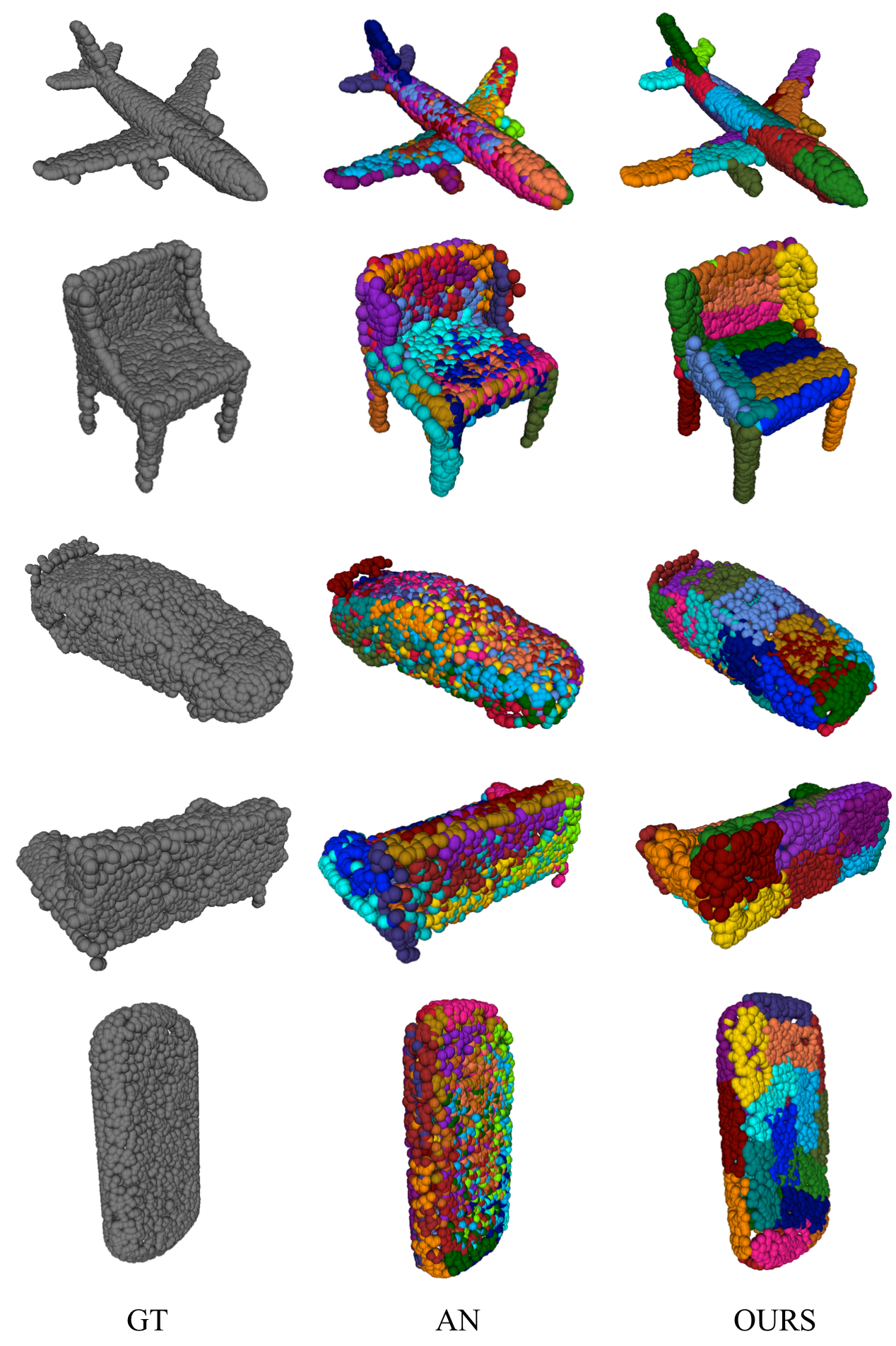}
	\caption{{\bf \OURS{} vs \AN{} trained for \PCAE{}} on the ShapeNet dataset. Each color denotes the points generated by one patch. Those generated by our approach are much better organized with far less overlap.}
	\label{fig:sn_qualitative}
\end{figure}

We report our results in Table \ref{tab:ours_vs_an_on_sn}. Note that \OURS{} delivers comparable CHD precision while achieving significantly less overlap and higher normals accuracy as quantified by metrics $\molap$ and $\mAE$. Fig. \ref{fig:growing_k_sn} depicts the mean overlap as a function of the neighborhood size threshold $t$ used to compute $\molap^{(t)}$. Our approach consistently reduces the overlap, as  illustrated 
by Fig.~\ref{fig:sn_qualitative}.

\begin{table}[t]
  \centering
  \caption{{\bf \OURS{} vs \AN{} trained for \PCAE{}.} Both models were trained individually on 5 object categories from the ShapeNet dataset. While CHD is comparable for both methods, \OURS{} delivers better normals and lower patch overlap.}
  \resizebox{0.47\textwidth}{!}{
  	\begingroup
  	\setlength{\tabcolsep}{2pt}
  	\renewcommand{\arraystretch}{0.8}
    \begin{tabular}{cccccccc}
    \hspace{-2mm}\textbf{obj.} & \textbf{method} & \textbf{CHD} & $\mathbf{\mAE}$ & \boldmath{}\textbf{$\molap^{(0.01)}$}\unboldmath{} & \boldmath{}\textbf{$\molap^{(0.05)}$}\unboldmath{} & \boldmath{}\textbf{$\molap^{(0.1)}$}\unboldmath{} & \textbf{$\mathbf{\mcol}$} \\
    \midrule
    \multirow{2}[2]{*}{plane} & AN    & \textbf{1.07} & 21.26 & 5.90  & 12.08 & 15.39 & 0.006 \\
          & \OURS{}  & 1.08  & \textbf{17.90} & \textbf{3.82} & \textbf{7.99} & \textbf{10.88} & \textbf{0.000} \\
    \midrule
    \multirow{2}[2]{*}{chair} & AN    & \textbf{2.79} & 24.49 & 5.30  & 9.45  & 12.12 & 0.011 \\
          & \OURS{}  & 2.82  & \textbf{23.06} & \textbf{2.85} & \textbf{5.78} & \textbf{8.09} & \textbf{0.000} \\
    \midrule
    \multirow{2}[2]{*}{car} & AN    & 4.68  & 18.08 & 4.61  & 9.07  & 12.50 & 0.011 \\
          & \OURS{}  & \textbf{3.34} & \textbf{17.75} & \textbf{2.50} & \textbf{4.85} & \textbf{7.26} & \textbf{0.000} \\
    \midrule
    \multirow{2}[2]{*}{couch} & AN    & \textbf{2.10} & 16.83 & 3.74  & 8.07  & 11.67 & \textbf{0.000} \\
          & \OURS{}  & 2.21  & \textbf{14.90} & \textbf{2.54} & \textbf{5.41} & \textbf{8.08} & \textbf{0.000} \\
    \midrule
    \multirow{2}[2]{*}{cellphone} & AN    & \textbf{1.80} & 10.29 & 6.51  & 13.65 & 16.79 & \textbf{0.000} \\
          & \OURS{}  & 1.82  & \textbf{9.64} & \textbf{2.75} & \textbf{6.13} & \textbf{8.69} & \textbf{0.000} \\
    \bottomrule
    \end{tabular}
	\endgroup
}	
  \label{tab:ours_vs_an_on_sn}
\end{table}

\parag{SVR on TDS.} 

We ran two separate experiments on the T-Shirt and the Cloth. We use $4$ patches for the former as the object represents a simple rectangular shape and $14$ for the latter as the T-Shirt is more complex. We trained \AN{} using the code provided by the authors.  For both \OURS{} and \AN{}, we used $8000$  points randomly sampled from the GT and the same number is predicted by the networks. We trained both \OURS{} and \AN{} until convergence.

\begin{figure}[htb]
	\centering
	\includegraphics[width=0.8\linewidth]{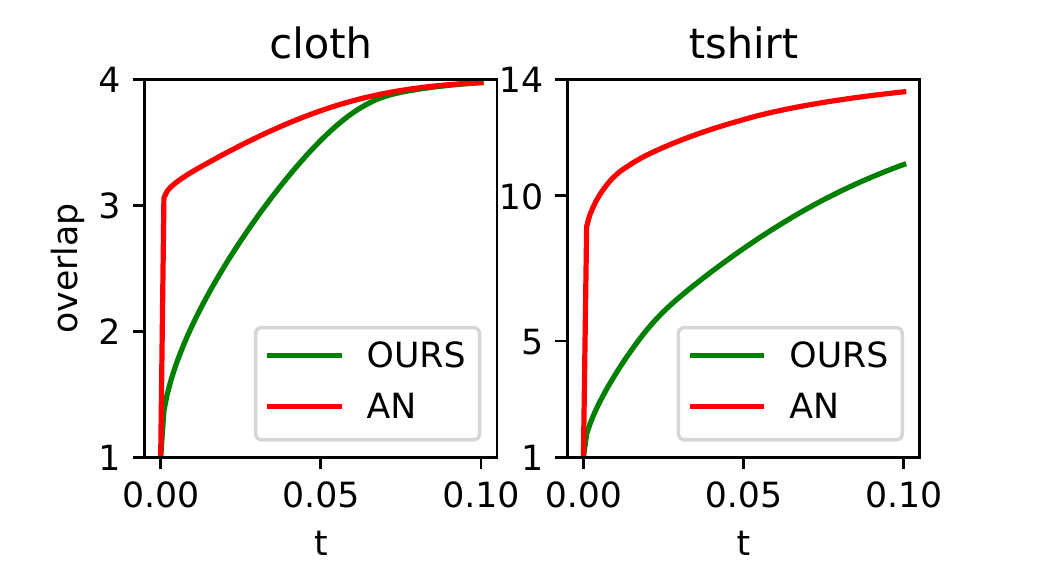}
	\caption{{\bf Patch overlap for \OURS{} and \AN{}} trained for \PCAE{} on the TDS dataset. We plot $\mcol^{(t)}$ as a function of $t$.}
	\label{fig:growing_k_tds}
\end{figure}

\begin{figure}[htb]
	\centering
	\includegraphics[width=0.99\linewidth]{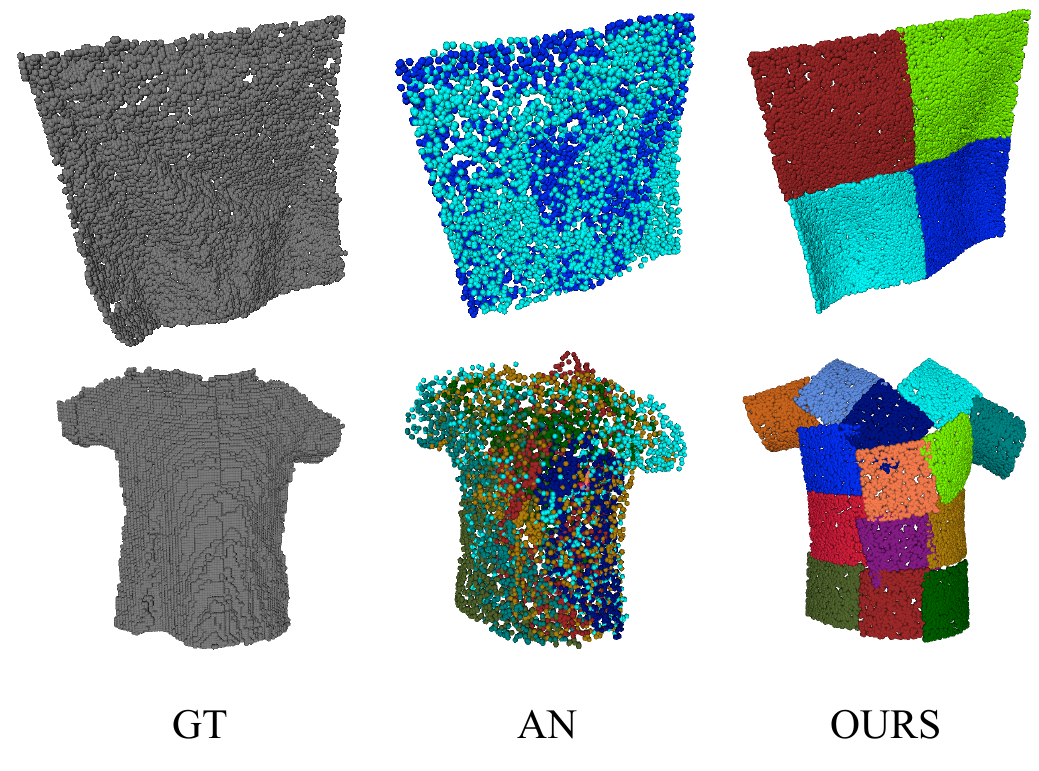}
	\caption{{\bf  \OURS{} vs \AN{} trained for \SVR{}} on the TDS dataset. Each color denotes the points generated by one patch. Those generated by our approach are much better organized with far less overlap and no collapsed patches.}
	\label{fig:tds_qualitative}
\end{figure}

\begin{figure}[htb]
	\centering
	\includegraphics[width=0.7\linewidth]{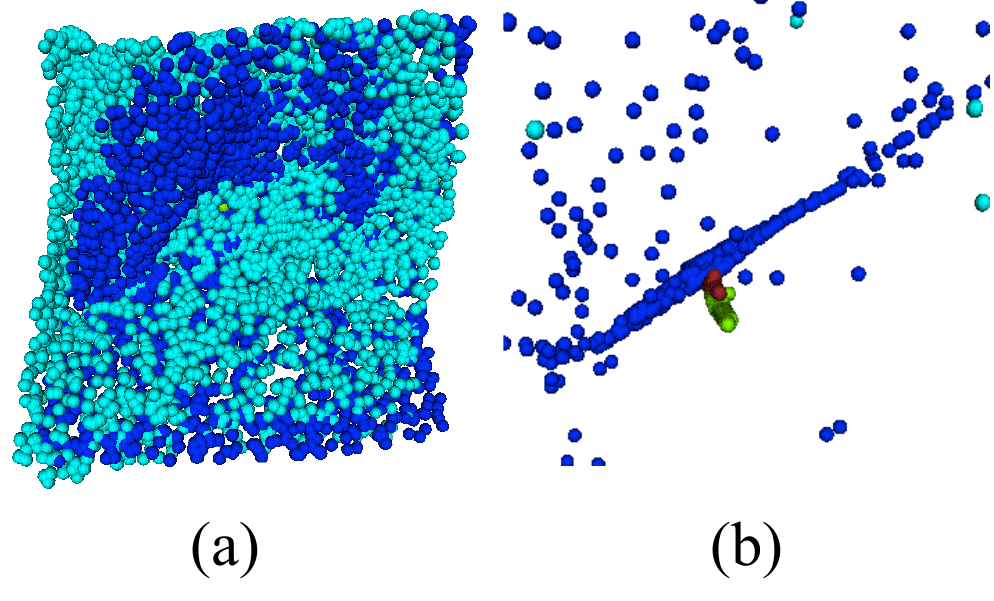}
	\caption{{\bf Partial collapse.} Even though the dark blue patch predicted by \AN{} seems to be well developed (a), a zoomed in view reveals a partial collapse (b).}
	\label{fig:partial_collapse}
\end{figure}

We report the results in Table~\ref{tab:ours_vs_an_on_tds}. The results show the same trends as in the previous example, with a similar accuracy in terms of CHD but a higher normal accuracy and much less overlap for \OURS{}. The qualitative results are depicted in Fig.~\ref{fig:tds_qualitative} and the amount of overlap is quantified in Fig.~\ref{fig:growing_k_tds}. Note that in this case, \AN{} suffers a number of patch collapses whereas \OURS{} does not, which means that if the normals and curvature were needed for future processing our approach would be the better option. Besides the obvious 0D point collapses, the predictions of \AN{} also
suffer less visible but equally harmful partial collapses as demonstrated in Fig.~\ref{fig:partial_collapse}.

\begin{table}[htbp]
  \centering
  \caption{{\bf \OURS{} vs \AN{} on \SVR{} for $\mathbf{2}$ objects from the TDS dataset.} As before CHD is comparable for both methods, but \OURS{} delivers better normals and less patch overlap.}
  \resizebox{0.47\textwidth}{!}{
  	\begingroup
  	\setlength{\tabcolsep}{2pt}
  	\renewcommand{\arraystretch}{0.8}
    \begin{tabular}{cccccccc}
    \textbf{object} & \textbf{method} & \textbf{CHD} & $\mathbf{\mAE}$ & \boldmath{}\textbf{$\molap^{(0.001)}$}\unboldmath{} & \boldmath{}\textbf{$\molap^{(0.005)}$}\unboldmath{} & \boldmath{}\textbf{$\molap^{(0.05)}$}\unboldmath{} & $\mathbf{\mcol}$ \\
    \midrule
    \multirow{2}[2]{*}{cloth} & AN    & \textbf{0.26} & 47.42 & 3.06  & 3.19  & 3.76  & 2 \\
          & \OURS{}  & 0.28  & \textbf{20.06} & \textbf{1.37} & \textbf{1.75} & \textbf{3.53} & \textbf{0} \\
    \midrule
    \multirow{2}[2]{*}{tshirt} & AN    & 0.35  & 42.12 & 8.95  & 10.03 & 12.64 & 7 \\
          & \OURS{}  & \textbf{0.31} & \textbf{20.52} & \textbf{1.80} & \textbf{2.89} & \textbf{8.22} & \textbf{0} \\
    \bottomrule
    \end{tabular}%
	\endgroup
}	
  \label{tab:ours_vs_an_on_tds}
\end{table}%

%% file: tex/conclusion.tex
% !TEX root = ../top.tex
% !TEX spellcheck = en-US

\section{Conclusion}
\label{sec:conclusion}

We have presented a novel and generic deep learning framework for 3D cloud point generation that makes it possible to compute analytically the differential properties of the surface the 3D points represent, without any need for post-processing. Our approach is inspired by the multi-patch approach of~\cite{Groueix18} and we have shown that we can use those differential properties during training to reduce the amount of patch overlap while delivering usable normals and curvatures, which the original approach does not do. 

In future work, we will incorporate this framework in end-to-end trainable networks that require the differential properties to exploit the image information and  to  perform tasks such as shape-from-shading or texture mapping.

%% file: tex/supplementary_arxiv.tex
%%%%%%%%%%%%%%%%%%%%%%%%%%%%%%%%%%%%%%%%%%%%%%%%%%%%%%%%%%%%%%%%%%%%%%%%%%%%%%%%%%%%%%%%%%%%%%%%%%%%
\section{Supplementarty Material}\label{sec:supplementary_material}

We provide more details on the training and evaluation of Single-View 3D Shape Reconstruction (\SVR{}) on the TDS dataset in Section~\ref{sec:etails_of_evaluation_of_svr_on_tds_dataset}; we perform an ablation study of the
components of the deformation loss $\Ldeform$ in Section~\ref{sec:deformation_loss_term_ablation_study};
and finally we analyze thoroughly  the deformation properties of the predicted patches in 
Section~\ref{sec:distortion_analysis}.

%%%%%%%%%%%%%%%%%%%%%%%%%%%%%%%%%%%%%%%%%%%%%%%%%%%%%%%%%%%%%%%%%%%%%%%%%%%%%%%%%%%%%%%%%%%%%%%%%%%%
\subsection{Training and Evaluation of SVR on TDS} \label{sec:etails_of_evaluation_of_svr_on_tds_dataset}

To evaluate the reconstruction quality of AN and \OURS{} for SVR on the TDS dataset, some preprocessing and postprocessing steps are necessary. 

The TDS dataset samples are centered around point 
$\cb = \begin{bmatrix}0 & 0 & 1.1\end{bmatrix}^{\top}$, which is out of reach of the 
activation function \textit{tanh} that AN uses in its last layer. Therefore, we translated all the data 
samples by $-\cb$.

In \cite{Bednarik18}, which introduced the TDS dataset, the authors align the predicted sample with its GT using Procrustes alignment~\cite{Stegmann02} before evaluating the reconstruction quality. Since we do not have  correspondences between the GT and predicted points, we used the Iterative Closest Point (ICP)~\cite{Besl92} algorithm to align the two point clouds. 
This allows rigid body transformations only.

%%%%%%%%%%%%%%%%%%%%%%%%%%%%%%%%%%%%%%%%%%%%%%%%%%%%%%%%%%%%%%%%%%%%%%%%%%%%%%%%%%%%%%%%%%%%%%%%%%%%
\subsection{Deformation Loss Term Ablation Study} \label{sec:deformation_loss_term_ablation_study}
We have seen that the deformation loss term defined as 
$\Ldeform = \alphE\LE + \alphG\LG + \alphskew\Lskew + \alphstr\Lstretch$ prevents the predicted 
patches from collapsing. Here we perform an ablation study of the individual components 
$\LE, \LG, \Lskew$ and $\Lstretch$ and show how each of them affects the resulting deformations 
that the patches undergo. 

We carry out all the experiments on \SVR{} using the cloth object from the TDS dataset and the same
training/validation/testing splits as before. We employ \OURS{} and the original loss function 
$\Lmain = \Lchd + \alphdef\Ldeform + \alpholap\Loverlap$ 
(with $\alphdef = 0.001$ and $\alpholap = 100$, as before). 

To identify the contributions of the 
components of $\Ldeform$, we switch them on or off by setting their corresponding hyperparameters 
$\alphE, \alphG, \alphskew$ and $\alphstr$ to either $0$ or $1$, and for each configuration we 
train \OURS{} from scratch until convergence. We list the individual configurations in 
Table~\ref{tab:ablation}.

\begin{table}[h]
  \centering
  \caption{{\bf Configurations of the ablation study.} The components of the
  $\Ldeform$ loss are either turned on or off using their corresponding hyperparameters.}
    \begin{tabular}{lcccc}
    Experiment & $\alphE$ & $\alphG$ & $\alphskew$ & $\alphstr$ \\
    \midrule
    free  & 0     & 0     & 0     & 0 \\
    no collapse & \textbf{1}     & \textbf{1}     & 0     & 0 \\
    no skew & \textbf{1}     & \textbf{1}     & \textbf{1}     & 0 \\
    no stretch & \textbf{1}     & \textbf{1}     & 0     & \textbf{1} \\
    full  & \textbf{1}     & \textbf{1}     & \textbf{1}     & \textbf{1} \\
    \end{tabular}%
  \label{tab:ablation}
\end{table}

\begin{figure*}[h]
	\centering
	\includegraphics[width=0.99\linewidth]{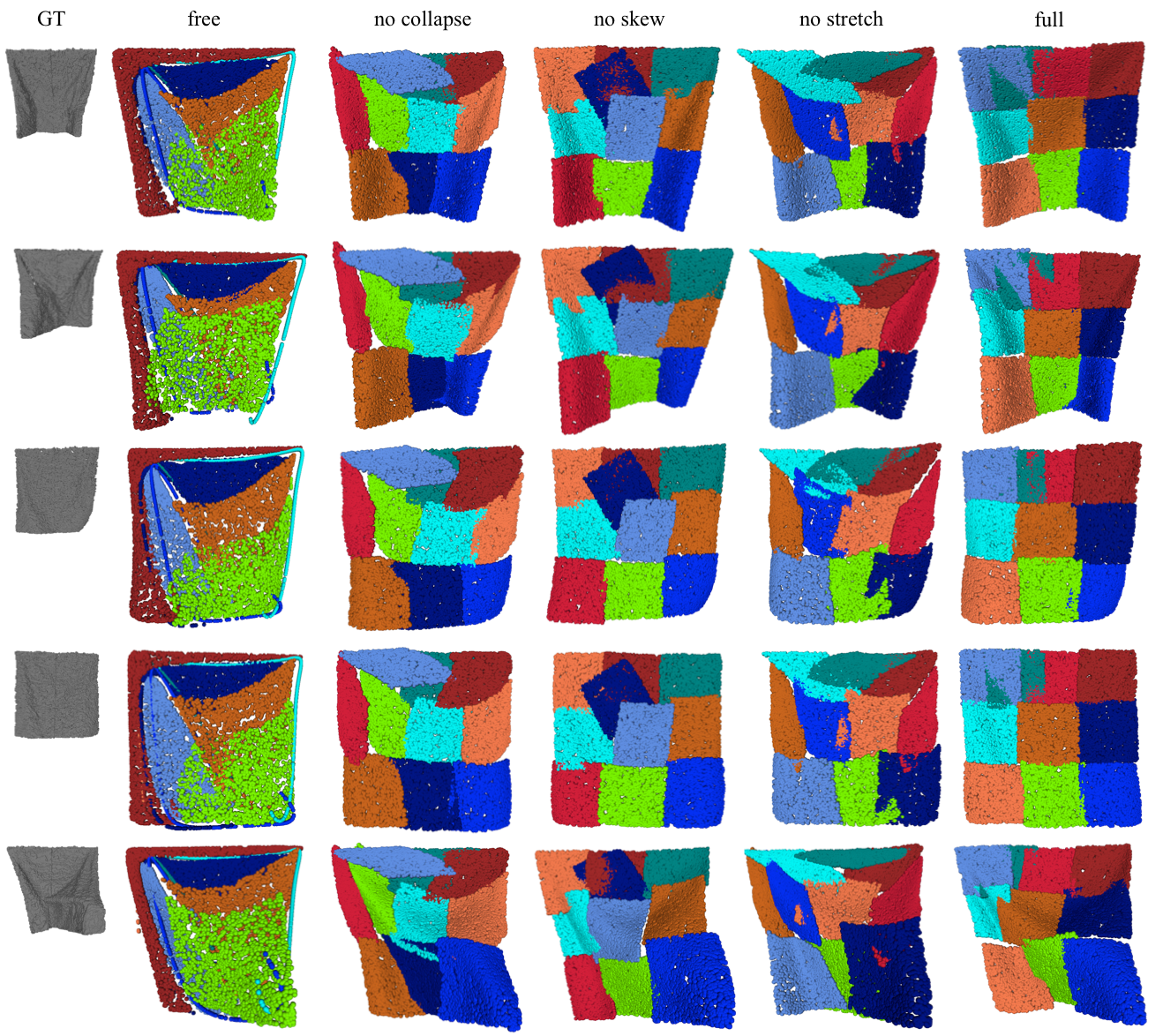}
	\caption{{\bf Qualitative results of the ablation study.} Each row depicts a
	randomly selected sample from the test set and each column corresponds to one experimental
	configuration. See the text for more details.}
	\label{fig:ablation}
\end{figure*}

Fig \ref{fig:ablation} depicts the qualitative results for all $5$ experiments on $5$ randomly
selected test samples. We discuss the individual cases below:

\parag{Free:} The $\Ldeform$ term is completely switched off, which results in high distortion 
mappings and many 0D point collapses and 1D line collapses.

\parag{No collapse:} We only turn on the components $\LE$ and $\LG$, which by design prevent any 
collapse and encourage the amount of stretching along either of the axes to be uniform across 
the whole area of a patch. However, the patches still tend to undergo significant stretch along 
one axis (light red patch) and/or display a high amount of skew (light blue and light orange patch).

\parag{No skew:} Adding the $\Lskew$ component to $\LE$ and $\LG$ (but leaving out $\Lstretch$) 
prevents the patches from skewing, resulting in strictly orthogonal rectangular shapes. However, 
the patches tend to stretch along one axis (light blue and light red patch). If skew is needed
to model the local geometry, the patches stay rectangular and rotate instead (dark blue patch).

\parag{No stretching:} Adding the $\Lstretch$ component to $\LE$ and $\LG$ (but leaving out
$\Lskew$) results in a configuration where the patches prefer to undergo severe skew 
(cyan and dark green patch), but preserve their edge lengths.

\parag{All:} Using the full $\Ldeform$ term, with all its components turned on, results
in strictly square patches with minimum skew or stretching.

%%%%%%%%%%%%%%%%%%%%%%%%%%%%%%%%%%%%%%%%%%%%%%%%%%%%%%%%%%%%%%%%%%%%%%%%%%%%%%%%%%%%%%%%%%%%%%%%%%%%
\subsection{Distortion Analysis} \label{sec:distortion_analysis}

In the previous section, we showed that the individual types of deformations that the patches may undergo----stretching, skewing and in extreme cases collapse----can be effectively controlled
by suitable combination of the components of the loss term $\Ldeform$. In this section, we present a different perspective on the distortions which the patches undergo. We focus on a texture mapping task where we show that using the $\Ldeform$ to train a network helps learn mappings with much less distortion. Furthermore, we inspect each patch individually and analyze how the distortion distributes
over its area.

\subsubsection{Regularity of the Patches}

We experiment on \PCAE{} using the ShapeNet dataset, on which we train AN and \OURS{} as in Section~\ref{sec:comparison_to_sota},
i.e., using the full loss function $\Lmain = \Lchd + \alphdef\Ldeform + \alpholap\Loverlap$ 
with $\alphdef = 0.001$ and $\alpholap = 100$ and with 
$\alphE = \alphG = \alphskew = 1, \alphstr = 0$. Furthermore, we train one more model, 
\textit{\OURS{}-{\bf \textit{strict}}}, which is the same as \OURS{} except that we set $\alphstr = 1$. In other words, 
\OURS{}-{\bf strict} uses the full $\Ldeform$ term where even stretching is penalized. 

To put things in perspective, when considering the ablation study of Section~\ref{sec:deformation_loss_term_ablation_study}, \AN{} corresponds to the \textit{free} configuration,
\OURS{} to the \textit{no skew} configuration and \OURS{}-{\bf strict} to the \textit{full} configuration.

\begin{figure*}[h]
	\centering
	\includegraphics[width=0.8\linewidth]{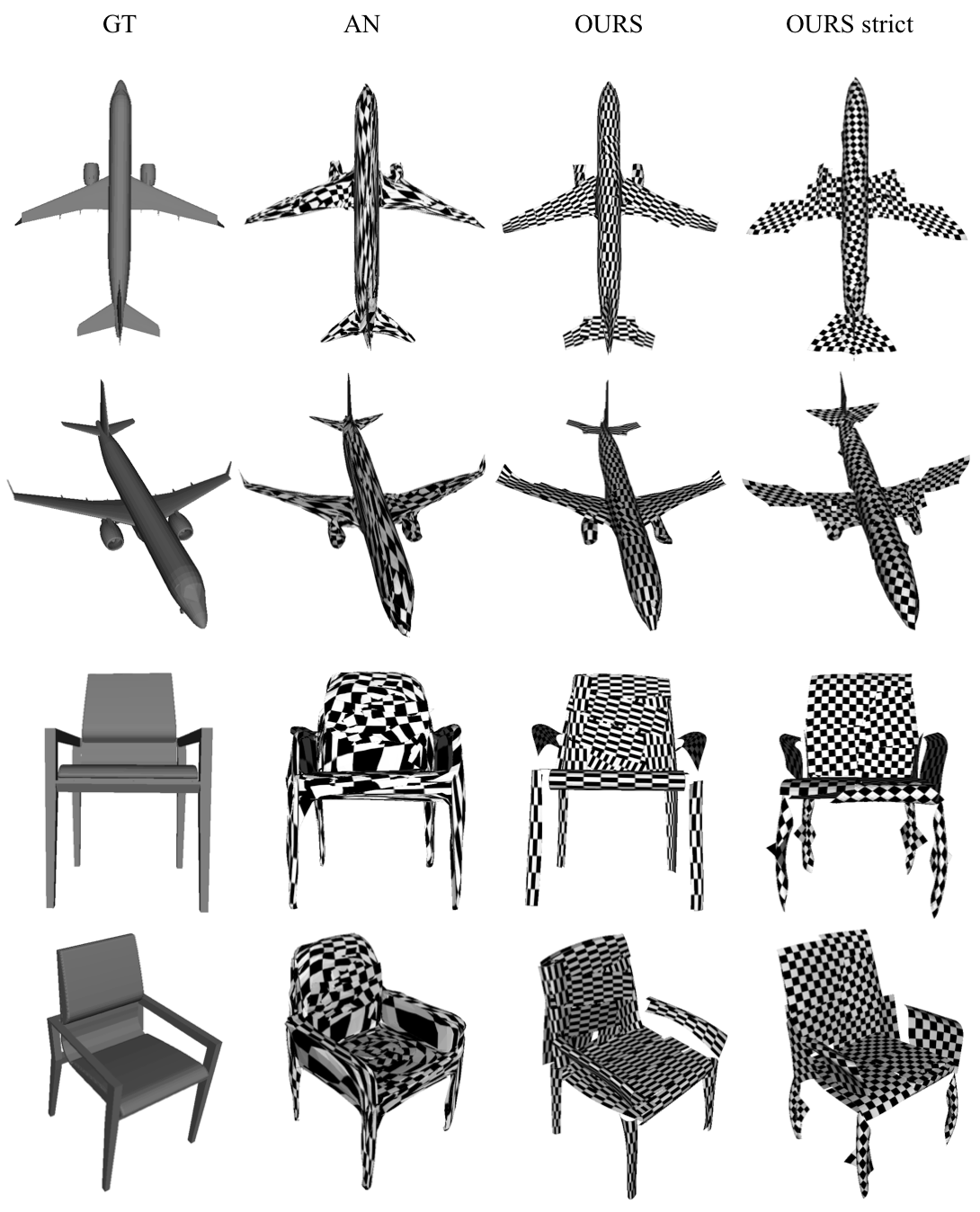}
	\caption{{\bf Qualitative results of ShapeNet objects plane and chair reconstruction by AN, 
			\OURS{} and \OURS{}-{\bf strict}.}}
	\label{fig:distortion_shapenet_1}
\end{figure*}

\begin{figure*}[h]
	\centering
	\includegraphics[width=0.8\linewidth]{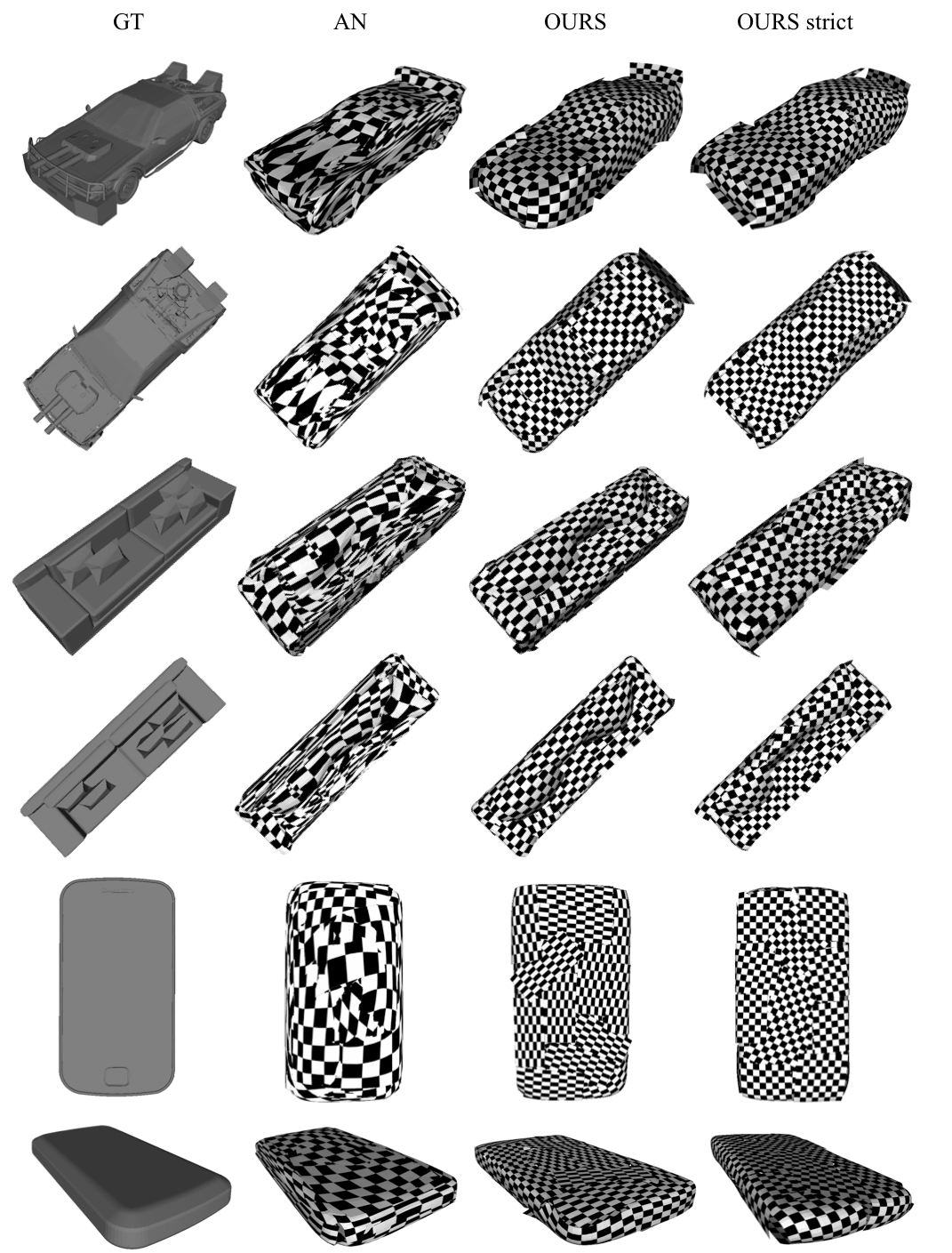}
	\caption{{\bf Qualitative results of ShapeNet objects car, couch and chair reconstruction 
			by AN, \OURS{} and \OURS{}-{\bf strict}.}}
	\label{fig:distortion_shapenet_2}
\end{figure*}

Figs.~\ref{fig:distortion_shapenet_1} and~\ref{fig:distortion_shapenet_2} depict 
qualitative reconstruction results for various objects from ShapeNet, where we map a 
regular checkerboard pattern texture to every patch. Note that while AN 
produces severely distorted patches, \OURS{} introduce a truly regular pattern elongated along
one axis (since stretching is not penalized) and \OURS{}-{\bf strict} delivers nearly isometric patches.

Note, however, the trade-off between the shape precision and regularity of the mapping (i.e., the
amount of distortion). When considering the two extremes, AN delivers much higher precision than 
\OURS{}-{\bf strict}. On the other hand, \OURS{} appears to be the best choice as it brings the best of both 
worlds --- it delivers high precision reconstructions while maintaining very low distortion mappings.

\subsubsection{Intra-patch Distortions}

To obtain more detailed insights into how the patches deform, we randomly select a test 
data sample from the ShapeNet plane object category and analyze the individual types of deformations
that each patch predicted by AN and \OURS{} undergoes. We are interested in $4$ quantities 
$\DE, \DG, \Dsk, \Dstr$, which are proportional to the components $\LE, \LG, \Lskew, \Lstretch$ of
the deformation loss term $\Ldeform$.

\begin{figure*}[h]
	\centering
	\includegraphics[width=0.99\linewidth]{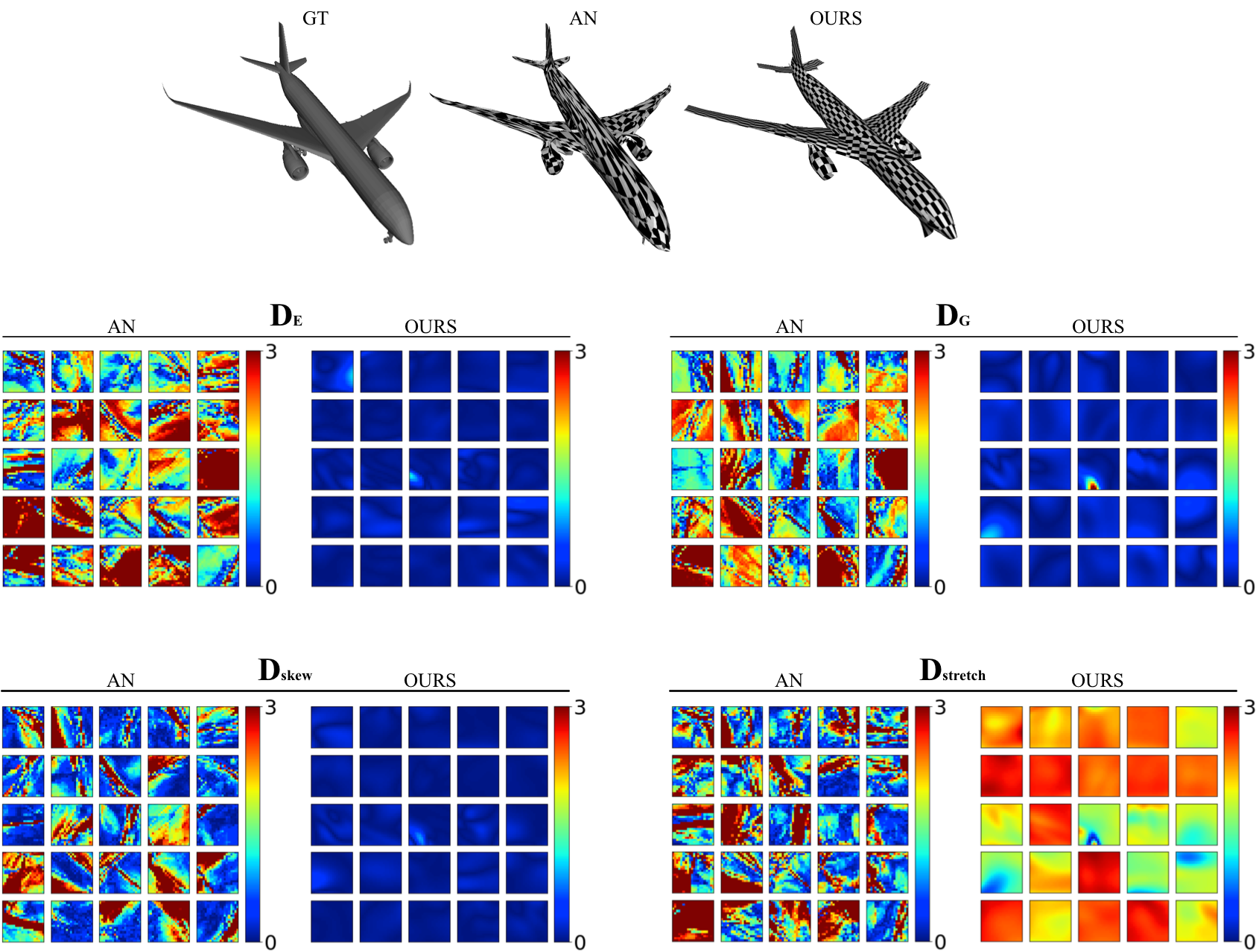}
	\caption{{\bf Spatial distribution of the quantities $\DE, \DG, \Dsk, \Dstr$ across all the 
			25 patches predicted by AN and \OURS{} for a single test data sample from ShapeNet dataset.}}
	\label{fig:distortion_patches}
\end{figure*}

Fig.~\ref{fig:distortion_patches} depicts the spatial distribution of the values coming from all 
these $4$ quantities over all $25$ patches predicted by AN and \OURS{}. Note that while the patches
predicted by AN are subject to all the deformation types and yield extremely high values, which change
abruptly throughout each predicted patch, the patches predicted by \OURS{} undergo very low distortions,
which are mostly constant throughout the patches. 

\begin{figure*}[h]
	\centering
	\includegraphics[width=0.45\linewidth]{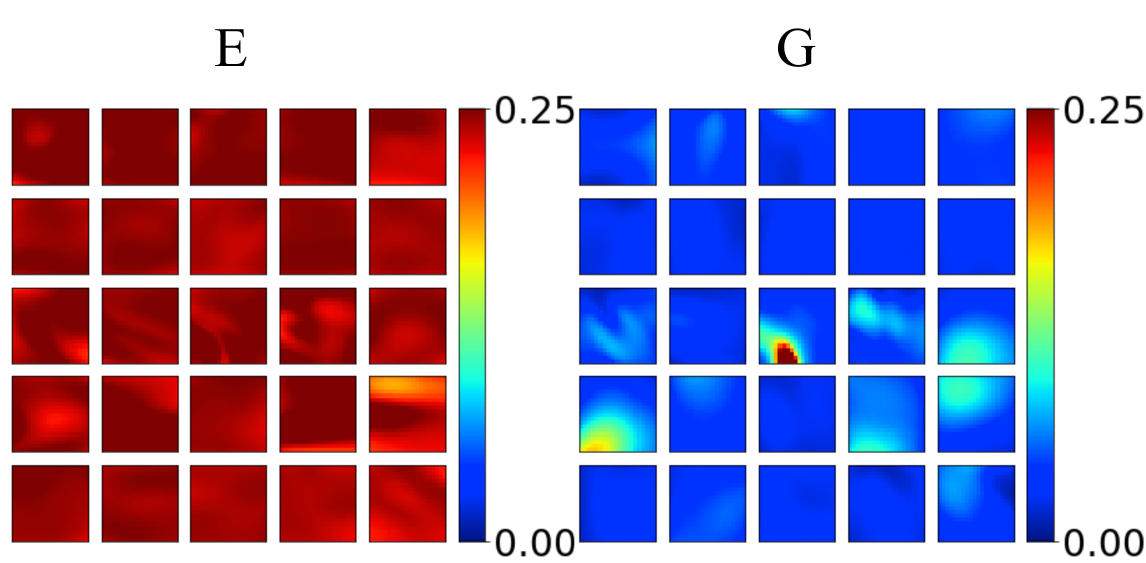}
	\caption{{\bf Spatial distribution of metric tensor $g$ quantities $E$ and $G$ over all the 
			25 patches predicted by \OURS{} on a single data sample from ShapeNet dataset.}}
	\label{fig:distortion_E_vs_G}
\end{figure*}

The exception is the $\Dstr$ quantity, which has high values for all the patches. This is due to the 
fact that \OURS{} does not penalize stretching. This can be seen in Fig.~\ref{fig:distortion_E_vs_G},
which depicts the distribution of the values of the terms $E$ and $G$ coming from the metrics tensor
$g = \begin{bmatrix}E & F \\ F & G\end{bmatrix}$ across all the patches predicted by \OURS{}. All the 
patches corresponding to $E$ yield high values while the ones corresponding to $G$ low values. This
means that the patches prefer to stretch only along the u-axis in the 2D parametric UV space 
(recall that $E = \normltwo{\frac{\partial f_{\wb}}{\partial u}}^{2}$).

